\documentclass[reqno]{amsart}
\usepackage[english]{babel}
\usepackage{lipsum}
\makeatletter
\usepackage{amsfonts}
\usepackage{amssymb}
\usepackage{centernot}
\usepackage{graphicx}
\usepackage{algorithm}
\usepackage{algorithmic}
\usepackage[algo2e]{algorithm2e}
\usepackage{etoolbox} %
\usepackage{fullpage}
\usepackage[pagewise]{lineno}
\usepackage{url}
\usepackage[dvipsnames, svgnames]{xcolor}
\usepackage[hidelinks]{hyperref}
\hypersetup{
  colorlinks   = true, 
  urlcolor     = Brown, 
  linkcolor    = DarkGreen, 
  citecolor   = DarkGreen 
}
\usepackage[normalem]{ulem}
\usepackage{caption}
\usepackage{subcaption}
\newcommand{\authorfootnotes}{\renewcommand\thefootnote{\@fnsymbol\c@footnote}}%
\makeatother
\newcommand*\linenomathpatch[1]{%
  \cspreto{#1}{\linenomath}%
  \cspreto{#1*}{\linenomath}%
  \csappto{end#1}{\endlinenomath}%
  \csappto{end#1*}{\endlinenomath}%
}

\linenomathpatch{equation}
\linenomathpatch{gather}
\linenomathpatch{multline}
\linenomathpatch{align}
\linenomathpatch{alignat}
\linenomathpatch{flalign}

\newtheorem{thm}{Theorem}[section]
\newtheorem{defn}[thm]{Definition}

\DeclareMathOperator*{\argmin}{arg\,min}
\DeclareMathOperator{\sgn}{sgn}

\newcommand{\bb}[1]{\mathbf{#1}}

\definecolor{darkgreen}{rgb}{.15,.55,0}

\definecolor{darkblue}{rgb}{0,0,0.7}

\colorlet{red-}{black}
\colorlet{blue-}{black}
\colorlet{orange-}{black}

\title{On the choice of the non-trainable internal weights in random feature maps for forecasting chaotic dynamical systems}

\begin{document}

\begin{abstract}
The computationally cheap machine learning architecture of random feature maps can be viewed as a single-layer feedforward network in which the weights of the hidden layer are random but fixed and only the outer weights are learned via linear regression. The internal weights are typically chosen from a prescribed distribution. The choice of the internal weights significantly impacts the accuracy of random feature maps. We address here the task of how to best select the internal weights. In particular, we consider the forecasting problem where random feature maps are used to learn a one-step propagator map for a dynamical system. We provide a computationally cheap hit-and-run algorithm to select {\em{good}} internal weights which lead to good forecasting skill. We show that the number of good features is the main factor controlling the forecasting skill of random feature maps and acts as an effective feature dimension. Lastly, we compare random feature maps with single-layer feedforward neural networks in which the internal weights are now learned using gradient descent. We find that random feature maps have superior forecasting capabilities whilst having several orders of magnitude lower computational cost.
\end{abstract}


\maketitle
{\normalsize
 \centering
  \authorfootnotes
  Pinak Mandal,\footnote{\thanks{pinak.mandal@sydney.edu.au, corresponding author}} \textsuperscript{1}
  Georg A. Gottwald, \footnote{\thanks{georg.gottwald@sydney.edu.au}}
  \textsuperscript{1} 
  Nicholas Cranch, \footnote{\thanks{ncra8185@uni.sydney.edu.au}}
  \textsuperscript{1} 

  \textsuperscript{1} The University of Sydney, NSW 2006,  Australia \par
 }

\section{Introduction}\label{sec-intro}
Estimation and prediction of the state of a dynamical system evolving in time is central to our understanding of the natural world and to controlling the engineered world. Often practitioners are tasked with such problems without the knowledge of the underlying governing dynamical system. In such scenarios a popular approach is to reconstruct the dynamical model from observations of the system \cite{meyer2000bayesian, abarbanel1993analysis,SmallEtAl02,BruntonEtAl16}. Predicting the future state of the system from these reconstructions is particularly challenging for chaotic dynamical systems. Chaotic dynamical systems cannot be accurately predicted beyond a finite time known as the predictability time due to their sensitive dependence on the initial conditions.

In recent times machine learning has achieved remarkable progress in learning surrogate models for dynamical systems from given data. Recurrent networks such as Long Short-Term Memory networks \cite{sherstinsky2020fundamentals, staudemeyer2019understanding} and gated recurrent units \cite{chung2014empirical} have been successfully applied in a plethora of time series prediction tasks \cite{fu2016using, cao2019financial, li2019ea}. These methods however often contain learnable parameters of the order of $\mathcal{O}(10^6)$, and require substantial fine tuning of hyperparameters and costly optimization strategies \cite{LiEtAl20}. An attractive alternative is provided by random feature maps \cite{RahimiRecht08,RahimiRecht08b,NelsenStuart21} and their extensions such as echo state networks and reservoir computers \cite{lukovsevivcius2009reservoir,lukovsevivcius2012practical,pathak2018model,nakajima2021reservoir}. These architectures can be viewed as a single-layer feedforward network in which the weights and biases of the hidden layer, the so-called internal parameters, are randomly selected before training and then are kept fixed. This renders the costly nonconvex optimization problem of neural networks to a simple linear least-square regression for the outer weights. The output of random feature maps and their extensions is hence a linear combination of a high-dimensional randomized basis. These methods have been shown to enjoy the universal approximation property, which states that in principle they can approximate any continuous function arbitrarily close \cite{rahimi2008uniform, barron1993universal,grigoryeva2018echo,gonon2021fading}.\\  

We focus here on classical random feature maps \cite{RahimiRecht08,RahimiRecht08b,NelsenStuart21} which have recently been shown to have excellent forecasting skill for chaotic dynamical systems \cite{GottwaldReich21,GottwaldReich21b}. The fact that random feature maps enjoy the universal approximation property does not provide practitioners with information on how to choose the internal parameters. The internal parameters are typically drawn from some prescribed distribution such as the uniform distribution on an interval or a Gaussian distribution. The forecasting capability of the learned surrogate map sensitively depends on the choice of the distribution \cite{GottwaldReich21}. To generate good internal parameters which lead to improved performance of random feature maps, several data-independent methods such as Monte Carlo and quadrature based algorithms as well as data-dependent methods such as leverage score based sampling and kernel learning have been proposed; for a detailed survey see \cite{liu2021random}. \textcolor{blue-}{Levine and Stuart \cite{LevineStuart22} performed Bayesian optimization to determine the internal weights together with other hyperparamters such as the regularization parameter.} Dunbar {\em{et al}} \cite{DunbarEtAl24} choose the distribution of the random weights from a parametric family. Its parameters are chosen to optimize a cost function motivated from Empirical Bayes arguments, with the optimization performed with derivative-free Ensemble Kalman inversion. Here we introduce a computationally cheap, non-parametric, optimization-free and data-driven method to draw internal parameters which lead to improved forecasting skill. We argue that good features, corresponding to good internal parameters, need to explore the expressive range of a given activation function. We consider here as an example the $\tanh$ activation function. To allow for good expressivity, good parameters should neither map the training data into the linear range of the activation function nor into the saturated range in which different inputs cannot be discerned. This leads us to a definition of {\em{good features}} corresponding to {\em{good internal parameters}}. We find that the set of good internal parameters is non-convex but can be expressed as a union of convex sets. To sample from a convex set we employ a hit-and-run algorithm \cite{smith1984efficient, zabinsky2013hit}. Hit-and-run algorithms are a class of Markov chain Monte Carlo samplers known for their fast mixing times in convex regions \cite{lovasz1999hit, lovasz2007geometry, laddha2023convergence}. In recent years, hit-and-run algorithms have also been analyzed for sampling nonconvex regions \cite{chandrasekaran2009thin, kiatsupaibul2011analysis, abbasi2017hit}.\\ 
The hit-and-run algorithms we develop allow us to generate any desired ratio of good features. We show in numerical experiments that the ratio of good features as defined by our criterion controls the forecasting capabilities of the learned surrogate map. Moreover, we illustrate the mechanism by which the least-square solution enhances good features and suppresses bad ones.\\
A secondary objective of our work is to demonstrate that a random feature map typically achieves superior forecasting skill when compared to a neural network of the same architecture, trained with gradient descent, while being several orders of magnitude cheaper computationally. We show that the bad performance of the single-layer feedforward network can be attributed to the optimization procedure not being constrained to the set of good internal parameters. This can potentially lead to new design and improved training schemes for more complex network architectures.\\

The outline of this paper is as follows. In Section~\ref{sec:problem} we describe the setup of data-driven surrogate maps for dynamical systems and how to assess their forecasting capabilities. Section~\ref{sec:RF} introduces random feature maps and illustrates how the choice of the internal weights affects the forecasting capabilities of the associated trained surrogate maps. Section~\ref{sec:ip} defines the set of good internal parameters and introduces hit-and-run algorithms to uniformly sample from this set. Section~\ref{sec:res} illustrates the effect of sampling from the good set of internal parameters on the forecasting skill and how the least-square training learns to distinguish good features associated with good parameters from those associated with internal parameters drawn from the complement of the good set. Section~\ref{sec:NN} compares random feature maps with single-layer feedforward networks in which the internal parameters are learned using backpropagation, and establishes that random feature maps with good parameters far outperform the single-layer feedforward neural network. \textcolor{red-}{Section~\ref{sec:longtime} shows that random feature maps with good internal parameters reproduce the long-time statistical behaviour of the underlying dynamical system more accurately than those without.} Finally, we conclude in Section~\ref{sec:conclude} with a summary of our results and possible future extensions.


\section{Dynamical setup}\label{sec:problem} We consider the forecasting problem for chaotic dynamical systems. Consider the following $D$-dimensional continuous-time dynamical system, 
\begin{align}
  \dot{\mathbf{u}} = \mathcal{F}(\mathbf{u}),
\label{eq:cts-system}
\end{align}
with initial data $\bb{u(0)} = \bb{u}_0$, which we observe at discrete times $t_n=n\Delta t$ for $n=0,1,\ldots, N$. We consider here the case when the full $D$-dimensional state is observed and observations are noise-free. For the treatment of noisy observations and partial observations see \cite{GottwaldReich21,GottwaldReich21b}. We view the dynamical system of these observations in terms of a discrete propagator map,
\begin{equation}
\begin{aligned}
    \mathbf{u}_{n+1} = \Psi_{\Delta t}(\mathbf{u}_{n}).
\end{aligned}\label{eq:dsc-system}
\end{equation}
The aim of data-driven modelling is to construct a surrogate map ${\hat{\Psi}}_{\Delta t}$ from the training data given by the observations that well approximates the true propagator map $\Psi_{\Delta t}$ of \eqref{eq:dsc-system}. In the following we denote variables associated with the surrogate map with a hat, and write the learned surrogate dynamical system as
\begin{align}
    \hat{\mathbf{u}}_{n+1} = {\hat{\Psi}}_{\Delta t}(\hat{\mathbf{u}}_n),
   \label{eq:dsc-surrogate}
\end{align}    
with initial data $\hat{\mathbf{u}}_0 = {\mathbf{u}}_0$. Throughout this work we use the $D=3$-dimensional Lorenz-63 system \cite{lorenz1963deterministic, lorenz1996predictability} with ${\bf{u}}=(x,y,z)$ and 
\begin{equation}
\begin{aligned}
    &\dot{x}=10(y-x),\\
    &\dot{y}=x(28-z)-y,\\
    &\dot{z}=xy-\frac{8}{3}z,
\end{aligned}\label{eq:L63}
\end{equation}
as the underlying continuous dynamical system \eqref{eq:cts-system}. The Lorenz-63 system is chaotic with a positive Lyapunov exponent of $\lambda_{\rm{max}}\approx0.91$ \cite{Tucker02}. We generate independent training and validation data, \textcolor{red-}{$\bb{u}_n$ and $\bb{u}_n^{\rm validation}$}, sampled at $\Delta t=0.02$ by randomly selecting independent initial conditions $\bb{u}_0$ \textcolor{red-}{and $\bb{u}_0^{\rm validation}$, respectively}. We discard an initial transient dynamics of $40$ time units to ensure that the dynamics has settled on the attractor.

To test the predictive capability of a surrogate model, we \textcolor{red-}{apply it to unseen validation data} and define the forecast time  $\tau_f$ associated with the surrogate model, 
\begin{align}
    \tau_f = \inf\left\{t_n \lambda_{\max}:\frac{\|\mathbf{\hat{u}}^{\rm validation}_n-\mathbf{u}^{\rm validation}_n\|_2^2}{\|\mathbf{u}^{\rm validation}_n\|_2^2}>\theta\right\}.
    \label{eq:def-tau_f}
\end{align} 
The forecast time is measured in Lyapunov time units and measures when the prediction of the learned surrogate map \eqref{eq:dsc-surrogate}, initialized at $\hat{\bb{u}}_0^{\rm validation}=\bb{u}_0^{\rm validation}$, significantly deviates from the true validation trajectory $\bb{u}_n^{\rm validation}$. We employ here an error threshold of $\theta=0.05$. 


\section{Random feature maps}
\label{sec:RF} 
We consider random feature maps to learn the surrogate map (\ref{eq:dsc-surrogate}) with
\begin{align}
    {\hat{\Psi}}_{\Delta t}(\mathbf{u}) = \mathbf{W}\sigma{(\mathbf{W}_{\rm in}\mathbf{u} + \mathbf{b_{\rm in}})},
    \label{eq:architecture}
\end{align}
where $\mathbf{u}$ is the $D$-dimensional state vector, $\mathbf{W}_{\rm in}\in \mathbb{R}^{D_r\times D}$ is the internal weight matrix, $\mathbf{b}_{\rm in}\in \mathbb{R}^{D_r}$ the internal bias and $\bb{W}\in\mathbb{R}^{D\times D_r}$ the outer weight matrix. The nonlinear activation function $\sigma$ is applied component wise and we choose here $\sigma=\tanh$. Random features are characterized by the internal weights $(\mathbf{W}_{\rm in},\mathbf{b}_{\rm in})$ being drawn before training from a prescribed distribution $p(w_{\rm in},b_{\rm in})$. The internal weights remain fixed and are not learned as it would be the case for a single-layer feedforward network which has the same architecture as in (\ref{eq:architecture}). Random feature maps can hence be seen as a linear combination of features which are the components of the $D_r$-dimensional random features vectors
\begin{align}
\boldsymbol{\phi} = \sigma{(\mathbf{W}_{\rm in}\mathbf{u} + \mathbf{b_{\rm in}})}.
\label{eq:feat}
\end{align}

The matrix $\mathbf{W}$, controlling the linear combinations of the features, is learned from training data $\bb{U}\in \mathbb{R}^{D\times N}$, the columns of which are the observations $\mathbf{u}_n,\;n=1,\ldots, N$ of the system \eqref{eq:dsc-system}. We do so by solving the following regularized optimization problem, 
\begin{align}
\mathbf{W}^* = \underset{\mathbf{W}}{\argmin}\;\mathcal{L}(\mathbf{W};\bb{U}, \textcolor{blue-}{\mathbf{W}_{\rm in},\mathbf{b_{\rm in}}}), 
\label{eq:lin-opt}
\end{align}
with loss function    
 \begin{align}   
\mathcal{L} (\mathbf{W};\bb{U}, \textcolor{blue-}{\mathbf{W}_{\rm in},\mathbf{b_{\rm in}}})  =\|\mathbf{W}\mathbf{\Phi}(\mathbf{U})-\mathbf{U}\|^2 + \beta\|\mathbf{W}\|^2.\label{eq:lin-opt-cost}
\end{align}
Here $\|\cdot\|$ denotes the Frobenius norm, $\beta>0$ is a regularization hyperparameter, and $\mathbf{\Phi}(\mathbf{U})$ is the feature matrix whose $n$-th column is given by,
\begin{align}
\boldsymbol{\phi}(\mathbf{u}_{n-1}) = \tanh{(\mathbf{W}_{\rm in}\mathbf{u}_{n-1} + \mathbf{b_{\rm in}})}.
\label{eq:cost}
\end{align}
The solution of the optimization problem \eqref{eq:lin-opt} is given explicitly by linear ridge regression as
\begin{align}
    \mathbf{W}^* = \mathbf{U}\mathbf{\Phi}(\mathbf{U})^\top(\mathbf{\Phi}(\mathbf{U})\mathbf{\Phi}(\mathbf{U})^\top+\beta\mathbf{I})^{-1}.
    \label{eq:sol-lr}
\end{align}
The low \textcolor{blue-}{training} cost of random feature maps makes them a very attractive architecture. \textcolor{blue-}{Training reduces to the determination of the $D_rD$ outer weights $\mathbf{W}^*$ by means of an explicit analytical formula once the inner weights have been set to their fixed values.} \textcolor{red-}{The total number of parameters for random feature maps is $D_r(2D+1)$ which scales linearly in the feature dimension $D_r$ unlike other architectures such as echo state networks \cite{jaeger2002tutorial, jaeger2004harnessing} and long short term memory (LSTM) networks  \cite{hochreiter1997long}.}


\subsection{The effect of the internal weights on the performance of random feature maps}
\label{sec:feature}
\textcolor{blue-}{Random feature maps enjoy the universal approximation property \cite{rahimi2008uniform, barron1993universal}. This means that for a given feature dimension $D_r$ there exist internal weights such that the random feature map approximates any continuous function. The approximation can get arbitrarily close for increasing feature dimension. The universal approximation property, however, does not guide practitioners on how to find the internal weights $(\mathbf{W}_{\rm in},\mathbf{b}_{\rm in})$ which allow for such an approximation.} The main objective of this paper is to sample the internal parameters in a way that increases the forecasting skill of the random feature maps \textcolor{red-}{when compared to the usually employed data-uninformed random draws from a specified distribution such as a Gaussian or a uniform distribution.}

\begin{figure}
    \centering
\includegraphics[scale=0.59]{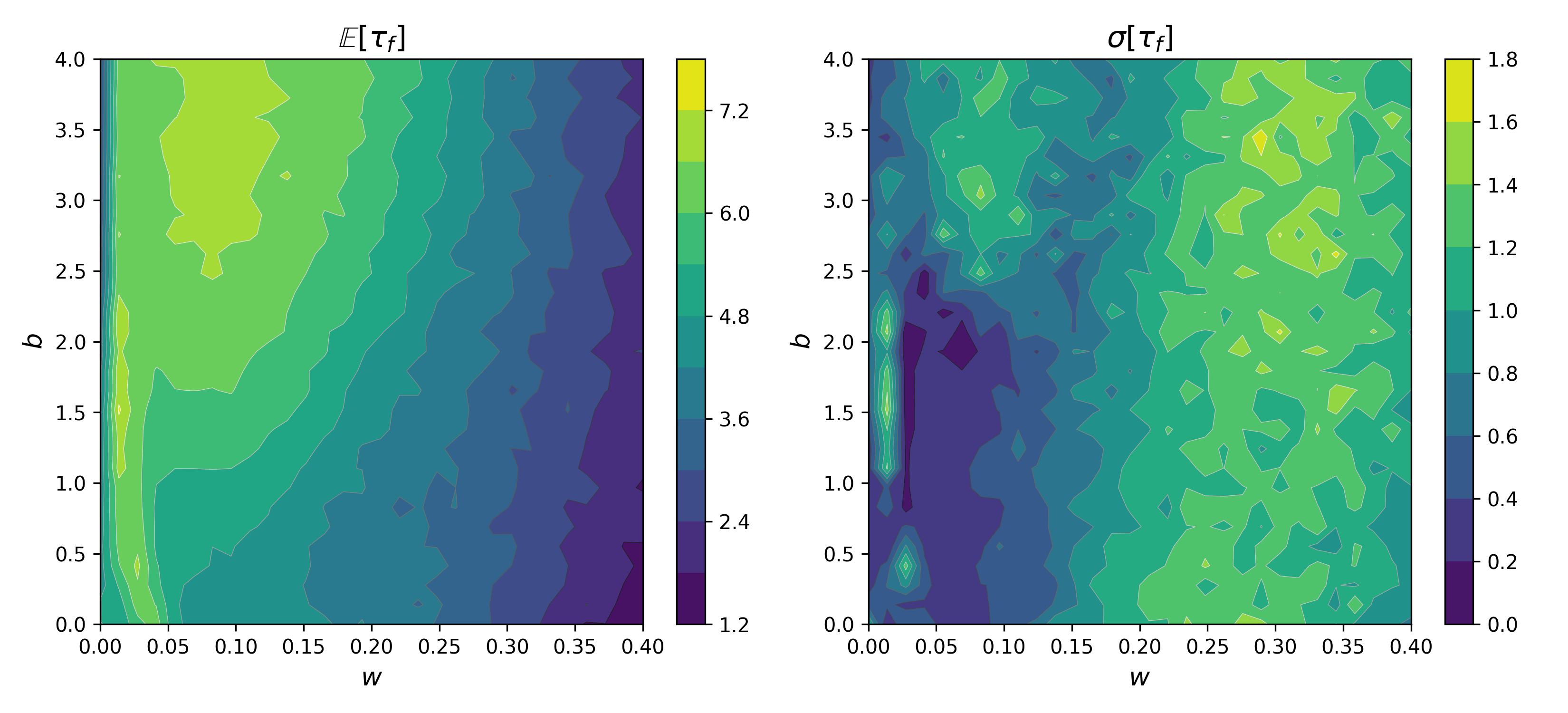}
    \caption{Contour plots of the mean and standard deviation of the forecast time $\tau_f$ computed using $\mathbf{W}_{\rm in},\mathbf{b}_{\rm in}$ sampled uniformly from intervals of variable size $[-w, w]$ and $[-b, b]$ respectively. Samples were drawn for grid points $(w, b)$ on a $30\times30$ regular grid over the domain $(0,0.4)\times(0, 4.0)$. Averages are taken over $M=100$ realizations per grid-point $(w, b)$, for a feature dimension $D_r=300$, training data length $N=20,000$ and regularization parameter $\beta=4\times10^{-5}$, using the same training and validation data for ecah realization.} 
    \label{fig:heat-tau_f}
\end{figure}

Indeed, the forecasting skill of a learned random feature map (\ref{eq:dsc-surrogate}) sensitively depends on the internal weights. To illustrate the effect of the hyperparameters $(\mathbf{W}_{\rm in},\mathbf{b}_{\rm in})$ on the forecast time $\tau_f$ we uniformly sample $\mathbf{W}_{\rm in}$ and $\mathbf{b}_{\rm in}$ from the intervals $[-w, w]$ and $[-b, b]$, respectively, with $(w, b)\in (0, w_{\rm max})\times (0, b_{\rm max})$. In particular, we use $30\times30$ regular grid points over $(0, w_{\rm max})\times (0, b_{\rm max})$ with $w_{\rm max}=0.4$ and $b_{\rm max}=4.0$, and probe the statistics by generating $M=100$ feature maps for each grid point, while keeping the training data and the validation data fixed for all realizations to focus on the effect of the internal weights. We fix the feature dimension at $D_r=300$ and the regularization parameter at $\beta=4\times 10^{-5}$. Figure~\ref{fig:heat-tau_f} shows a contour plot of the mean and the standard deviation of the forecast time $\tau_f$ over the domain of the internal weights. We can clearly see that certain regions in the hyperparameter space are associated with good performance with mean forecast times $\tau_f > 4$ while other regions produce poor mean forecast times. Moreover, regions in the hyperparameter space corresponding to high mean forecast times $\tau_f$ may have large variance. 

\begin{figure}[!htp]
    \centering
\includegraphics[scale=0.5]{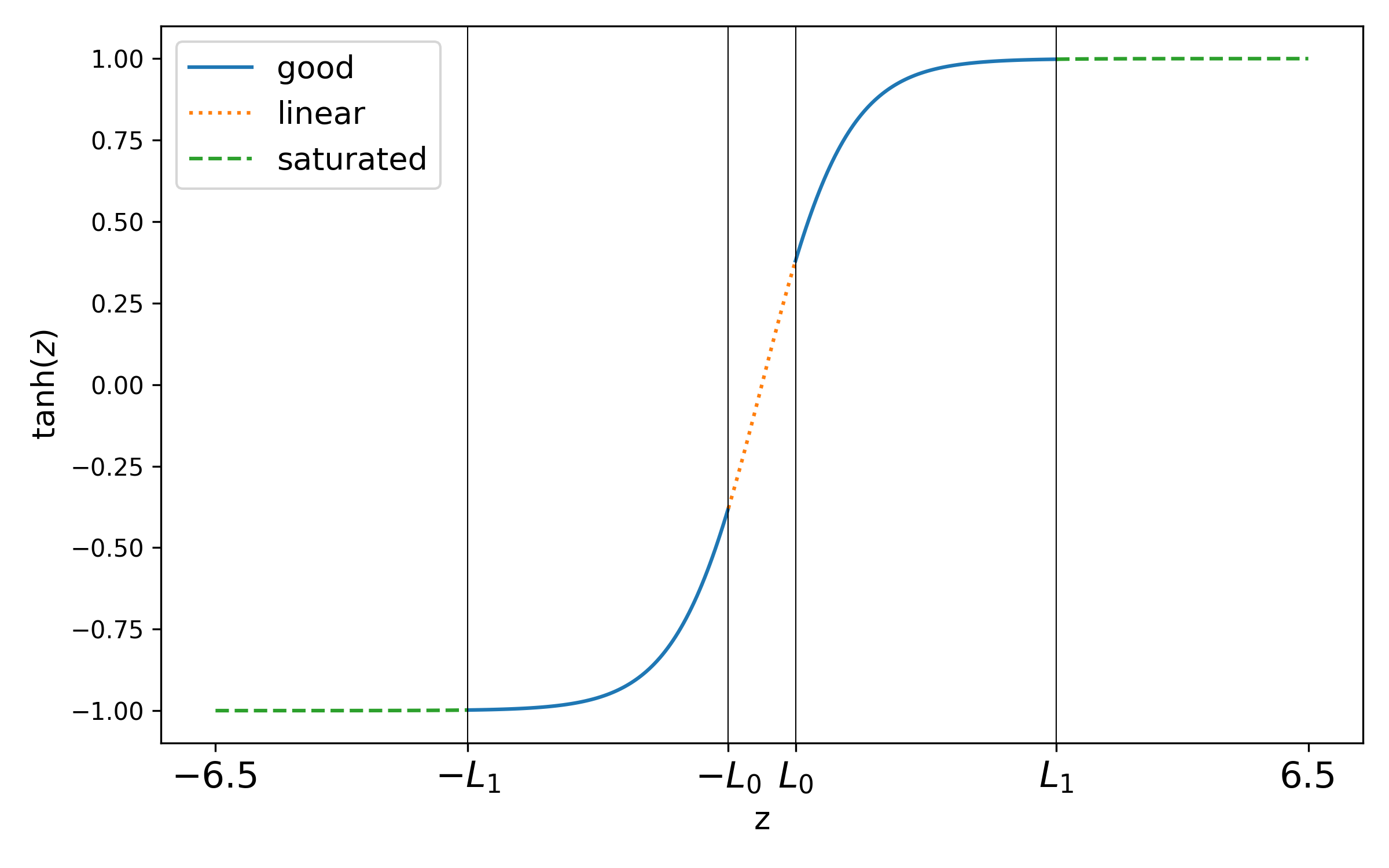}
    \caption{Domain and range of features produced by a $\tanh$-activation function  with $L_0=0.4$ and $L_1=3.5$, leading to linear, saturated and good features.}
    \label{fig:tanh}
\end{figure}

Ideally, we would like parameters which have both, high mean forecast time and low variance so that the performance is not dependent on the particular training data used. It is clear that if the internal weights $(\mathbf{W}_{\rm in},\mathbf{b}_{\rm in})$ are chosen sufficiently small, the associated features (\ref{eq:feat}) are essentially linear with $\boldsymbol{\phi} \approx \mathbf{W}_{\rm in}\mathbf{u} + \mathbf{b_{\rm in}}$ for all input data $\bb{u}$. This would reduce the random feature maps to linear models which are known to be incapable of modelling nonlinear chaotic dynamical systems \cite{GauthierEtAl21,Bollt21}. On the other extreme, for sufficiently large internal weights a $\tanh$-activation function saturates, and one obtains $\boldsymbol{\phi} \approx \pm 1$ independent of the input data $\bb{u}$, severely decreasing the expressivity of the random feature map. This suggests that one should choose internal weights which sample the $\tanh$-activation function in its nonlinear non-saturated range. This is illustrated in Figure~\ref{fig:tanh}. We shall call features \textit{linear}, if for all data $\bb{u}$ the argument of the $\tanh$-activation function lies within the interval centred around the origin in $[-L_0,L_0]$. Those features obtained by the $\tanh$-activation function that are approximately $\pm 1$ for all input data $\bf{u}$, i.e. where the arguments of the $\tanh$-activation function lie in either of two unbounded sets $(-\infty, -L_1],[L_1, +\infty)$, we label \textit{saturated} features. Those features which for all input data are neither linear nor saturated, i.e. for which the argument of the $\tanh$-activation function lies in either of the two intervals $(-L_1,-L_0)$ or $(L_0,L_1)$, are labelled {\textit{good}} features. We use $L_0=0.4$ and $L_1=3.5$ to define good, linear and saturated features throughout this paper.

\begin{figure}[!htp]
    \centering
\includegraphics[scale=0.5]{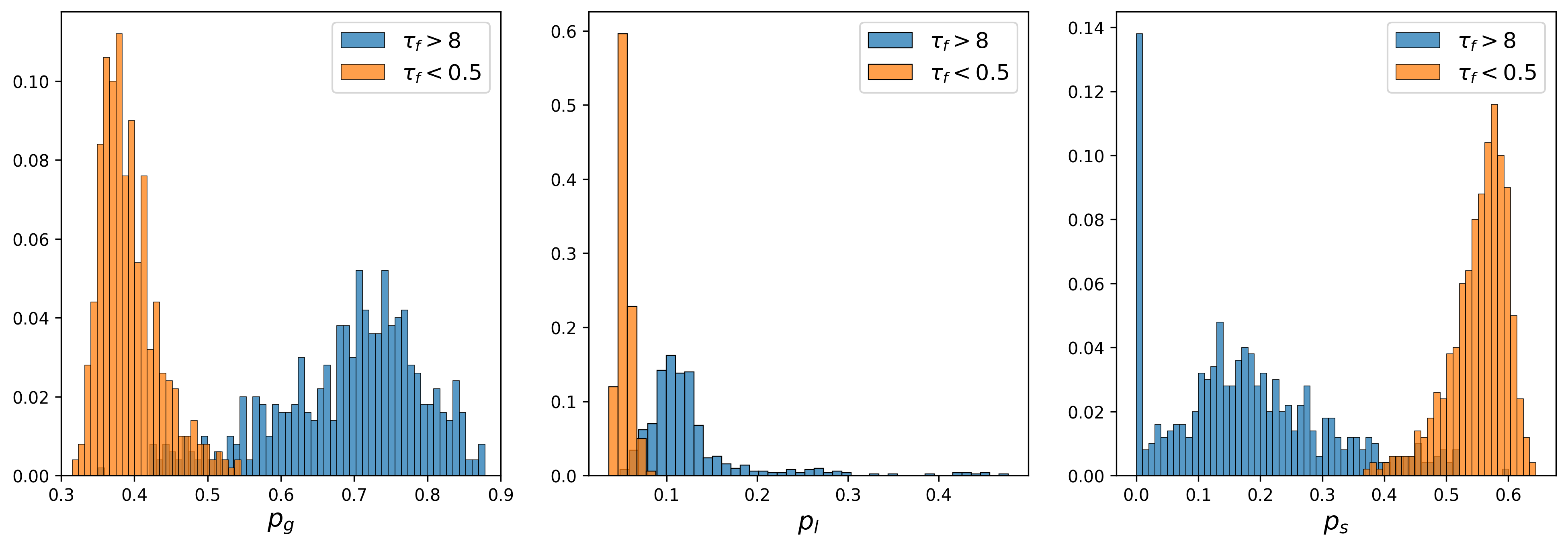}
\caption{\textcolor{blue-}{Empirical histograms of average fractions of good, linear and saturated features, $p_g$, $p_l$ and $p_s$, respectively, for random feature maps corresponding to two groups: large forecast times with $\tau_f>8$ and low forecast times with $\tau_f<0.5$. The random feature maps are the same as those used in Figure~\ref{fig:heat-tau_f}. Each group contains $500$ samples and the histograms depict the probability of having a certain value of the respective fractions in each group.}}
    \label{fig:saturated-hist}
\end{figure}

\textcolor{blue-}{We now illustrate the detrimental effect of linear and saturated features on the forecasting skill of random features. From the random feature maps that were used in Figure~\ref{fig:heat-tau_f}, we select those that lead to particularly large forecast times $\tau_f>8$ and those that lead to particularly low forecast times $\tau_f<0.5$. For each of those feature maps we determine the fraction of features that correspond to (a) the linear region $[-L_0,L_0]$, (b) the saturated region $(-\infty, -L_1] \cup [L_1, +\infty)$, and (c) the good region $(-L_1,-L_0)\cup (L_0,L_1)$, averaged over input data $\bb{u}_n$. We denote these fractions by $p_l$, $p_s$ and $p_g$, respectively. We consider $5,000$ randomly selected data points ${\bf{u}}_n$ on the attractor of the Lorenz-63 system to estimate these fractions. For each group we randomly select $500$ samples from the $90,000$ random feature maps used in Figure~\ref{fig:heat-tau_f}. In Figure~\ref{fig:saturated-hist} we show the histograms of the fractions $p_g$, $p_l$ and $p_s$ for these two groups. The group with low forecast times $\tau_f<0.5$ contains a significantly higher fraction of linear or saturated features compared to the group with large forecast times $\tau_f>8$. Conversely, the group with large forecast times $\tau_f>8$ contains a significantly higher fraction of good features compared to the group with low forecast times $\tau_f<0.5$. This confirms our hypothesis that good forecast skill is associated with an abundance of good features and a lack of linear and saturated features. We remark that the pronounced peak at $p_s=0$ is a sampling effect: when sampling uniformly from the grid $(0, w_{\rm max})\times (0, b_{\rm max})$ with $w_{\rm max}=0.4$ and $b_{\rm max}=4.0$, it is much more likely to draw parameters which correspond to non-saturated features. Such random feature map samples are much more likely to have higher forecast times and hence are concentrated entirely in the $\tau_f>8$ group.}

\begin{figure}[!htp]
    \centering
\includegraphics[scale=0.6]{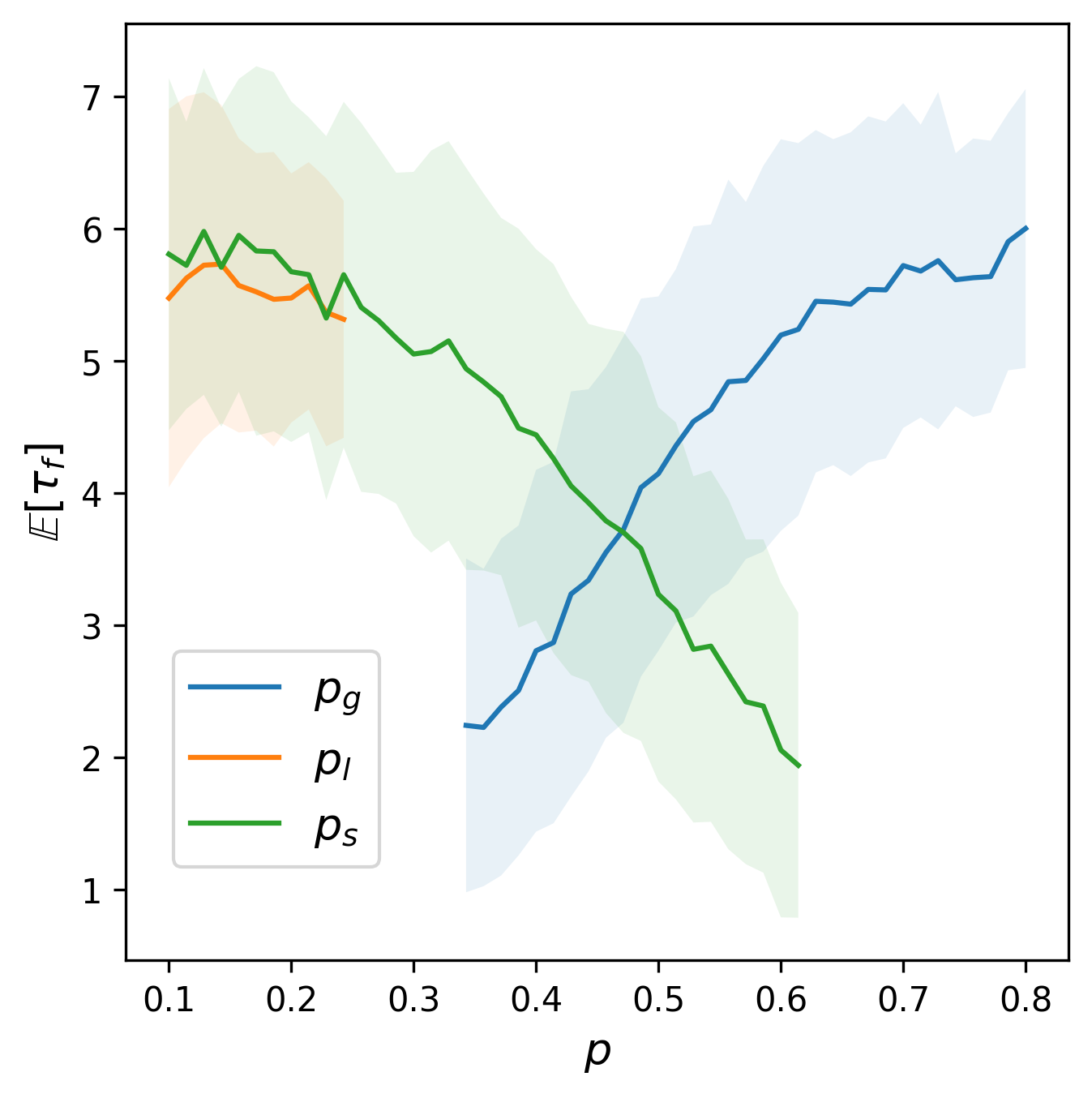}
\caption{\textcolor{red-}{Mean forecast time $\mathbb{E}[\tau_f]$ as a function of the fraction $p$ of good, linear and saturated features, respectively. The random feature maps are the same as those used in Figure~\ref{fig:heat-tau_f}.  The mean forecast times are computed over bins $[p-\Delta p, p+\Delta p]$ with $\Delta p=0.001$. The shaded region delineates one standard deviation from the mean. We only report on bins which contain more than $100$ samples.}}
    \label{fig:tauf_F}
\end{figure}

\textcolor{red-}{Figure~\ref{fig:tauf_F} shows the mean forecast time $\mathbb{E}[\tau_f]$ as a function of the fractions of good, linear and saturated features, for the same random feature maps as shown in Figure~\ref{fig:saturated-hist}. It is clearly seen that the forecast skill, as measured by the mean forecast time, monotonically improves for increasing number of good features. Similarly, the forecast skill monotonically degrades with increasing number of saturated features. The effect of the linear features is less obvious for the uniformly initialized random feature maps of Figure~\ref{fig:heat-tau_f}. However, as we will see later (cf. Figure~\ref{fig:tau_f-gls}), linear and saturated features have the same negative effect on the mean forecast time.}

In the following Section we develop a computationally cheap algorithm to sample from the set of good weights, and show in Section~\ref{sec:res} how this increases the forecasting skill of random feature maps.

%


\section{How to sample good internal weights}
\label{sec:ip}
We would like our random feature maps to produce good features $\boldsymbol{\phi}(\bb{u})=\mathbf{W}_{\rm in}\mathbf{u}+\mathbf{b}_{\rm in}$ by restricting $(\mathbf{W}_{\rm in}, \mathbf{b}_{\rm in})$ to be neither linear nor saturated for all training data $\mathbf{u}_n$. To that end, we select $(\mathbf{W}_{\rm in}, \mathbf{b}_{\rm in})$ such that 
\begin{align}
    L_0<|\mathbf{W}_{\rm in}\mathbf{u}_n+\mathbf{b}_{\rm in}|<L_1,\qquad\forall\;n=0,1,\ldots N. 
    \label{eq:ineq-good}
\end{align}
The lower bound $L_0$ controls the linear features and the upper bound $L_1$ controls the saturated features (cf. Figure~\ref{fig:tanh}). Note that \eqref{eq:ineq-good} is a vector inequality and is equivalent to $D_r$ scalar inequalities. Denoting the $i$-th row of  $\mathbf{W}_{{\rm in}}$ with $\bb{w}_i^{\rm in}$ and the $i$-th entry of $\bb{b}_{\rm in}$ with $b_i^{\rm in}$, for each $i\in\{1,2,\ldots D_r\}$ we require 
\begin{align}
    L_0<|\bb{w}_{i}^{\rm in}\cdot\mathbf{u}_n+b_i^{\rm in}|<L_1,\qquad\forall\;n=0,1,\ldots N. \label{eq:row-ineq-good}
\end{align}
{\defn We call the $i$-th row $(\mathbf{w}_i^{\rm in},b_i^{\rm in})$ of the internal parameters $(\bb{W}_{\rm in},\bb{b}_{\rm in})$ \textit{good} if it satisfies \eqref{eq:row-ineq-good}. Similarly, we call $(\mathbf{w}_i^{\rm in},b_i^{\rm in})$ \textit{linear} if 
\begin{align}
    |\bb{w}_{i}^{\rm in}\cdot\mathbf{u}_n+b_i^{\rm in}|\le L_0,\qquad\forall\;n=0,1,\ldots N, 
    \label{eq:row-ineq-linear}
\end{align}
and we call $(\mathbf{w}_i^{\rm in},b_i^{\rm in})$  \textit{saturated} if 
\begin{align}
    |\bb{w}_{i}^{\rm in}\cdot\mathbf{u}_n+b_i^{\rm in}|\ge L_1,\qquad\forall\;n=0,1,\ldots N.
   \label{eq:row-ineq-saturated}
\end{align}
For a streamlined discussion we call the $i$-th column of the outer weight matrix $\bb{W}$, good if the associated $i$-th row of the matrix of internal weights $(\bb{W}_{\rm in},\bb{b}_{\rm in})$ is good and so on. 
} 

This categorization of rows of the internal parameters is useful for investigating the effects of different realizations of the random feature map on its forecasting skill. Note that this is not an exhaustive classification since there exist rows that satisfy different inequalities for different observations ${\mathbf{u}}_n$ and do not satisfy \eqref{eq:row-ineq-good} for the whole data set $\mathbf{U}$. Although not considered here, such \textit{mixed} rows may be an interesting topic for further exploration. \textcolor{red-}{The distinction into ``good" (nonlinear, non-saturated) and ``bad" (linear or saturated) parameters only pertains to the task of longest possible sequential forecasting. For example, for the task of predicting the state for only a single step, all parameters, good or bad, perform equally well.} 

We denote the set of good internal weights satisfying \eqref{eq:ineq-good} by $\Omega_g$. The solution set $\Omega_g$ is not convex, but can be written as the disjoint union of two convex sets with
\begin{align}
    \Omega_g = S_-\cup S_+,\label{eq:set-good}
\end{align}
where
\begin{align}
    S_- \,& = \,\{(\bb{w},b)\in\mathbb R^{D+1}:-L_1<\mathbf{w}\cdot\mathbf{u}_n+b<-L_0\;\;\forall\;n=0,1,\ldots,N\},\\
    S_+ \,& = \, \{(\bb{w},b)\in\mathbb R^{D+1}:+L_0<\mathbf{w}\cdot\mathbf{u}_n+b<+L_1\;\;\forall\;n=0,1,\ldots,N\}.
\end{align}
Since the convex subsets are reflections of each other with
\begin{align}
    S_-=-S_+,
\label{eq:reflection}
\end{align}
it suffices to sample from only one of these convex sets and then uniformly sample the sign of the internal weights to sample from $\Omega_g$. Hence, the sampling problem is effectively a convex problem. Analogously, we define $\Omega_l$ and $\Omega_s$ to be the solution sets to the problems \eqref{eq:row-ineq-linear} and \eqref{eq:row-ineq-saturated} respectively, and again sampling these sets are convex problems. \textcolor{blue-}{We will see that the data-informed constraint for good parameters \eqref{eq:ineq-good} allows for sufficient variability in the features (cf. Section~\ref{sec:performance}).}

We present, in the next two subsections, algorithms to effectively sample from \textcolor{blue-}{the good set $\Omega_{g}$ to enhance the forecasting capabilities of random feature maps, as well as from the sets $\Omega_{l}$ and $\Omega_{s}$ to illustrate their effect on random feature maps in Section~\ref{sec:res}}. A naive choice of sampling algorithm would be to uniformly sample from the $D$-dimensional hypercube with the $2^D$ corners defined by the extremal training data points, and checking the inequality (\ref{eq:ineq-good}), if we want to sample from $\Omega_g$, let's say. This, however, is computationally very costly as typically the solution set only occupies a small region within that hypercube. 

\textcolor{blue-}{Our aim is to sample good parameters. There is no {\em{a priori}} reason to sample {\em{uniformly}} from the convex set of good parameters. However, as we will see, uniform sampling lends itself to an efficient implementation. Furthermore, we stress that sampling uniformly from the parameter space does not imply uniform sampling in the feature space.}

We consider a standard hit-and-run algorithm sampling from $\Omega_g$ in Section~\ref{sec:hit-run} and then present a faster, more efficient hit-and-run algorithm to sample from an equivalent restricted solution set in Section~\ref{sec:restrict}.


\subsection{Standard hit-and-run sampling of good internal parameters}
\label{sec:hit-run}
We now describe a computationally cheap and easy to implement numerical algorithm to uniformly sample from the solution sets \textcolor{blue-}{$\Omega_{g}, \Omega_{l}$ and $\Omega_{s}$}. We shall employ hit-and-run algorithms \cite{smith1984efficient, zabinsky2013hit}. To uniformly sample a set $\Omega$ with hit-and-run, one starts from a feasible point inside the set, considers the line through that point in a randomly chosen direction, and then randomly picks a point on the intersection of that line and the set $\Omega$ as a new point. This process is then repeated to generate further samples. For convex sets the hit-and-run samples converge to uniform samples in total variation distance. The convergence depends polynomially on the number of iterations and dimension with the polynomial dependence on dimension being of low order~\cite{lovasz1999hit, abbasi2017hit, lovasz2004hit}. This and the fast mixing properties make hit-and-run algorithms an attractive method to uniformly sample from \textcolor{blue-}{$\Omega_{g}, \Omega_{l}$ and $\Omega_{s}$}.

We sample the augmented internal weight matrix $(\mathbf{W}_{\rm in}, \mathbf{b}_{\rm in})$ row by row. Each sample lies then in a $D+1$-dimensional search space for $(\mathbf{w}_i^{\rm in},b_i^{\rm in})$. Due to (\ref{eq:reflection}) it suffices to sample from $S_+$. In order to perform hit-and-run, given a point, we need to efficiently determine if it lies in $S_+$. Focusing on a convex conical subset of $S_+$, it turns out that we can determine if a point belongs to $S_+$ by checking just two inequalities. Define the convex cone
\begin{align}
    V(\mathbf{s}, b)=\{(\mathbf{w},b):\sgn(\mathbf{w}_i)\in\{\mathbf{s}_i, 0\}\;\;\forall\;i=1,2,\ldots,D\},
\end{align}
where $\bb{s}$ is a $D$-dimensional sign vector with entries $\pm 1$ labelling the $2^D$ corners of a $D$-dimensional hypercube. To control the range of the training data set, we further define the vectors $\mathbf{x}_{\mp}(\bb{s}) \in\mathbb R^D$ as
\begin{equation}
\begin{aligned}
    \mathbf{x}_{-,i}(\bb{s}) = \begin{cases}
        &\underset{1\le n\le N}{\min}\; \mathbf{u}_{n,i},\qquad\text{if } \mathbf{s}_i=1\\
        &\underset{1\le n\le N}{\max}\; \mathbf{u}_{n,i},\qquad\text{otherwise}
        \end{cases}\\
    \mathbf{x}_{+,i}(\bb{s}) = \begin{cases}
        &\underset{1\le n\le N}{\max}\; \mathbf{u}_{n,i},\qquad\text{if } \mathbf{s}_i=1\\
        &\underset{1\le n\le N}{\min}\; \mathbf{u}_{n,i},\qquad\text{otherwise},
    \end{cases}
\end{aligned}\label{eq:xpm}
\end{equation}
where $\bb{u}_{n, i}$ is the $i$-th entry of the $n$-th training data point. Now for $(\bb{w},b)\in V(\bb{s}, b)$ we have, 
\begin{equation}
\begin{aligned}
    &\underset{1\le n\le N}{\max}\;\;({\mathbf{w}\cdot \bb{u}_n+b}) \le \mathbf{w\cdot x_+(\bb{s})}+ b,\\ 
\text{ and}\\
    &\underset{1\le n\le N}{\min}\;\;({\mathbf{w}\cdot \bb{u}_n+b}) \ge \mathbf{w\cdot x_-(\bb{s})}+ b.
\end{aligned}
\end{equation}
Therefore, for $(\bb{w},b)\in V(\sgn(\bb{w}),b)$, we have $(\bb{w},b)\in S_+$ if
\begin{equation}
\begin{aligned}
    &\mathbf{w\cdot x_-(\sgn(\bb{w}))}+ b > L_0,\\\text{ and }\\
    &\mathbf{w\cdot x_+(\sgn(\bb{w}))}+ b < L_1
\end{aligned}\label{eq:feasibility} 
\end{equation}
The feasibility inequalities \eqref{eq:feasibility} simply check if the internal weights $(\bb{w},b)$ map the training data into the smallest $D$-dimensional hypercube that contains the training data. 
 
To initialize the hit-and-run algorithm with a feasible point we choose $(\bb{w},b) = (\mathbf{0}, b_0) \in S_{+}$ for $b_0\in(L_0,L_1)$. To determine the line segments inside $S_+$ we use bisection together with the feasibility criterion \eqref{eq:feasibility}. The hit-and-run algorithm requires a few iterations to ensure that the samples become independent of the initial feasible point $(\bb{w},b) = (\mathbf{0}, b_0)$. 

We summarize this hit-and-run algorithm for randomly generating uniform samples from $\Omega_g$ in Algorithm~\ref{algo:hr}.
\begin{algorithm}[!htp]
\caption{Standard hit-and-run sampling for a good row}
\label{algo:hr}
\begin{algorithmic}[1]
\STATE Input: data $\bb{U}$. 
\STATE Choose number of decorrelation iterations $K\in\mathbb{N}$ and $L_0, L_1\in\mathbb{R}_{>0}$. 
\STATE Sample $b$ uniformly from $(L_0, L_1)$.
\STATE $\bb{w}\leftarrow\bb{0}$.
\STATE $k\leftarrow0$.
\WHILE{$k< K$}
    \STATE Randomly select a unit vector $\bb{d}\in\mathbb{R}^{D+1}$.
    \STATE Determine $A=$ the maximal line segment passing through $(\bb{w}, b)$ along direction $\bb{d}$ and contained within $S_+$, using the feasibility criterion \eqref{eq:feasibility} and the bisection method. 
    \STATE Uniformly sample $(\bb{w}', b')$ from $A$.
    \STATE $(\bb{w}, b)\leftarrow (\bb{w}', b')$
    \STATE $k\leftarrow k+1$.
\ENDWHILE
\STATE Uniformly sample a scalar $z$ from $\{-1, 1\}$ to determine which set to sample from, $S_-$ or $S_+$.
\IF{$z=1$}
    \STATE $(\bb{w},b)$ is our final good row sample. 
\ELSE
    \STATE $(-\bb{w},-b)$ is our final good row sample. 
\ENDIF
\end{algorithmic}
\end{algorithm}


\subsection{One-shot hit-and-run sampling}
\label{sec:restrict}
We now present a reduced hit-and-run algorithm which operates on a smaller $D$-dimensional search space and does not require computationally costly bisection. This algorithm, which we will coin one-shot hit-and-run algorithm, produces independent samples without the need for sufficiently many iterations to ensure decorrelation from the fixed initial feasible point.

\begin{figure}[!htp]
    \centering
    \includegraphics[scale=0.86]{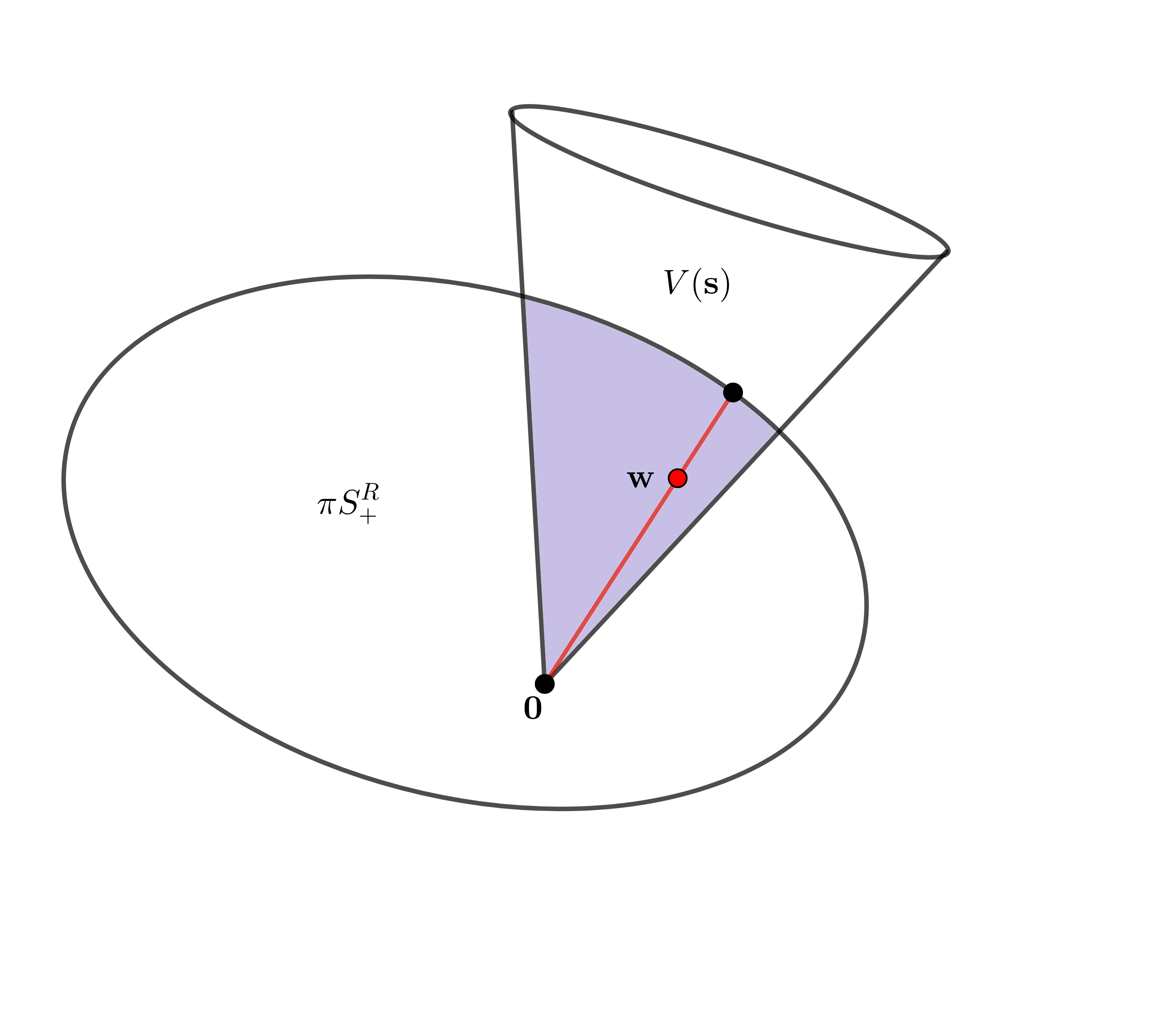}
    \caption{Schematic of the one-shot hit-and-run Algorithm~\ref{algo:hr-D}. The weight point $\bb{0}$ is always an interior point of $\pi S_+^R$ and the cone $V(\bb{s})$ is a $D$-dimensional orthant. The set $\pi S_+^R$ is drawn as bounded here, but it may be unbounded depending on the training data $\bb{u}_n$.}
    \label{fig:rhr}
\end{figure}

To generate good (or linear or saturated) random feature maps one can restrict the solution set \textcolor{blue-}{$\Omega_{g}$ (or $\Omega_{l}$ or $\Omega_{s}$)} by first sampling $b$ appropriately and then sampling $\bb{w}$ from a $D$-dimensional search space. For ease of presentation, we describe the algorithm for sampling from the good set $\Omega_g$. We sample $b$ uniformly from the interval $(L_0, L_1)$. The weights $\bb{w}$ are then sampled from the restricted solution set $\Omega_g^R=S_+^R\cup S_-^R$ with
\begin{align}
S_-^R&= \{(\bb{w},b)\in S_-:-L_1<b<-L_0\},\\S_+^R&= \{(\bb{w},b)\in S_+:+L_0<b<+L_1\}.
\label{eq:b-restriction}
\end{align}
Since $S_-^R=-S_+^R$, sampling from the nonconvex set $\Omega_g^R$ can again be done by sampling from the convex set $S_+^R$ and then multiplying the sample with $1$ or $-1$ uniformly randomly. This restriction allows us to perform hit-and-run sampling on a $D$-dimensional random convex set instead of a $(D+1)$-dimensional convex set. Note that fixing $b$ is akin to shrinking the search space from $S_+$ to $\pi S_+^R$ where $\pi$ is the canonical projection: $\pi(\bb{w}, b)=\bb{w}$. Also note that we can partition $\pi S_+^R$ according to
\begin{align}
    \pi S_+^R=\bigsqcup_{\bb{s}\in \{-1, 1\}^D}(\pi S_+^R\cap V(\bb{s})),\label{eq:pi-S+}
\end{align}
where we use $\bigsqcup$ to denote almost disjoint union, and $V(\bb{s})=\pi V(\bb{s},b)$ are $D$-dimensional orthants. Let us randomly select a sign vector $\bb{s}\in\{-1, 1\}^D$ or equivalently pick the random convex subset $\pi S_+^R\cap V(\bb{s})$. Randomly choosing the sign vector or the corresponding convex subset is tantamount to assigning signs randomly to the entries of $\bb{w}$. In order to uniformly sample this conical subset we can pick a random direction $\bb{d}$ in the cone $V(\bb{s})$, determine the maximal line segment starting at the origin parallel to $\bb{d}$ that is contained in $\pi S_+^R\cap V(\bb{s})$ and uniformly sample a point on this line segment. Figure~\ref{fig:rhr} shows a schematic for this one-shot hit-and-run algorithm.
Since $\bb{x}_\pm(\sgn(\bb{w}))$ is constant for all $\bb{w}\in V(\bb{s})$, we can analytically determine the maximal line segment without having to resort to bisection. Moreover, the special structure of the cone lets us sample with a single iteration unlike the standard hit-and-run Algorithm~\ref{algo:hr}. Thus the computation of the line segment in the solution set and the final sampling both happen in one shot and therefore the one-shot hit-and-run is much faster than its standard counterpart given by Algorithm~\ref{algo:hr}. 

Algorithm~\ref{algo:hr-D} summarizes the one-shot hit-and-run sampling of a good row. Note that, depending on the training data $\bb{U}$, it is possible for $\pi S_+^R$ to be unbounded, which is why $+\infty$ appears in the algorithm. We can extend the notion of restriction to the coordinates of $\bb{w}$ as well by restricting the intervals where we are allowed to sample them from, which is akin to regularizing parameters in machine learning \cite{kukavcka2017regularization, krogh1991simple, goodfellow2016deep} but we do not consider such algorithms here.  Obvious modifications of Algorithm~\ref{algo:hr} and Algorithm~\ref{algo:hr-D} let us sample linear and saturated rows which we refrain from describing here to avoid repetition.
%
\begin{algorithm}[!htp]
\caption{One-shot hit-and-run sampling for a good row}
\label{algo:hr-D}
\begin{algorithmic}[1]
\STATE Input: data $\bb{U}$. 
\STATE Choose $L_0, L_1\in\mathbb{R}_{>0}$.
\STATE Sample $b$ uniformly from $(L_0, L_1)$.
\STATE Select the sign vector $\bb{s}$ by uniformly generating $D$ samples from $\{-1, 1\}$.
\STATE Randomly select a unit vector $\bb{d}\in V(\bb{s})$. 
    \STATE $a_0\leftarrow0$.
    \STATE $a_1\leftarrow\inf\left(\left\{\frac{L_0-b}{\bb{d}\cdot \bb{x}_-(\bb{s})},\frac{L_1-b}{\bb{d}\cdot \bb{x}_+(\bb{s})}\right\}\cap(\mathbb R_{>0}\cup\{+\infty\})\right)$ with the convention $\inf\varnothing=+\infty$.
    \STATE Sample $a$ uniformly from $(a_0, a_1)$.
\STATE Uniformly sample a scalar $z$ from $\{-1, 1\}$ to determine which set to sample from, $S_-$ or $S_+$.
\IF{$z=1$}
    \STATE $(a\bb{d},b)$ is our final good row sample. 
\ELSE
    \STATE $(-a\bb{d},-b)$ is our final good row sample. 
\ENDIF
\end{algorithmic}
\end{algorithm}


\subsection{Performance \textcolor{blue-}{and comparison} of the hit-and-run algorithms}
\label{sec:performance}
\begin{figure}[!htp]
    \centering
    \includegraphics[scale=0.53]{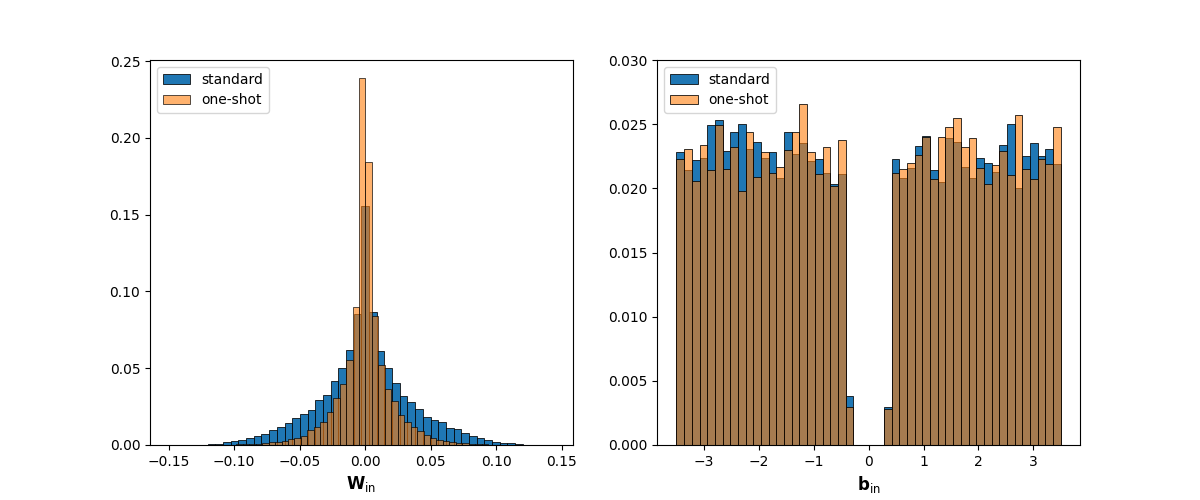}
    \caption{Empirical histograms for samples generated using standard and one-shot hit-and-run Algorithms~\ref{algo:hr} and \ref{algo:hr-D}, respectively. The left panel shows the distributions of the entries of $\bb{W}_{\rm in}$ and the right panel shows the distributions of the entries of $\bb{b}_{\rm in}$. For each algorithm $10,000$ rows of internal parameters were generated. The standard hit-and-run Algorithm~\ref{algo:hr} used $K=10$ decorrelation iterations.}
    \label{fig:hr-params}
\end{figure}

The two hit-and-run Algorithms~\ref{algo:hr} and \ref{algo:hr-D} are designed to uniformly sample from the \textcolor{blue-}{$D+1$-dimensional set of good rows $\Omega_g$. This does not imply that the individual entries of $(\bb{W}_{\rm in}, \bb{b}_{\rm in})$ are uniformly distributed.} The distributions for the entries of weights and biases sampled from the two hit-and-run algorithms are shown in Figure~\ref{fig:hr-params}. For the standard hit-and-run Algorithm~\ref{algo:hr} it was found that $K=10$ decorrelation steps were sufficient and results were very similar for $K=100$ iterations. It is clearly seen that the distributions are far from being the usually employed uniform or Gaussian distribution. The distributions are very similar for both algorithms. In particular, the standard Algorithm~\ref{algo:hr} exhibits the same lack of biases $b$ with small absolute value, as the one-shot hit-and-run Algorithm~\ref{algo:hr-D}. 

Whereas the one-shot hit-and-run algorithm excludes biases $b$ with absolute values smaller than $L_0$ by design, this may seem surprising for the standard hit-and-run algorithm. This can be explained as follows. For $0<b<L_0$ and $(\bb{w}, b)\in S_+$ we require that $\bb{w}\cdot\bb{u}$ lies in the positive interval $(L_0-b, L_1-b)$ for all training data $\bb{u}$. Since the directions of the vectors ${\bb{u}}$ in the training data are typically distributed over some range, $\bb{w}\cdot\bb{u}$ is typically not sign-definite for all data points ${\bb{u}}$. This implies that for all parameters in $\Omega_g$ we typically have $|b|>L_0$; a similar argument shows that typically $|b|<L_1$.  Hence, for typical data $\bb{u}$, we have $\Omega_g^R = \Omega_g$ and the search space of the one-shot hit-and-run Algorithm~\ref{algo:hr-D} is the same as that of the standard Algorithm~\ref{algo:hr}.  

For the weights and biases which were obtained by sampling uniformly from an interval as in  Figure~\ref{fig:heat-tau_f}, we checked that the weights corresponding to high forecasting skill indeed all satisfy our criterion of being good rows \eqref{eq:ineq-good}. This highlights the advantage of our non-parametric sampling over sampling strategies involving a set of parametrized distributions.\\ 

\begin{figure}[!htp]
    \centering
\includegraphics[scale=0.6]{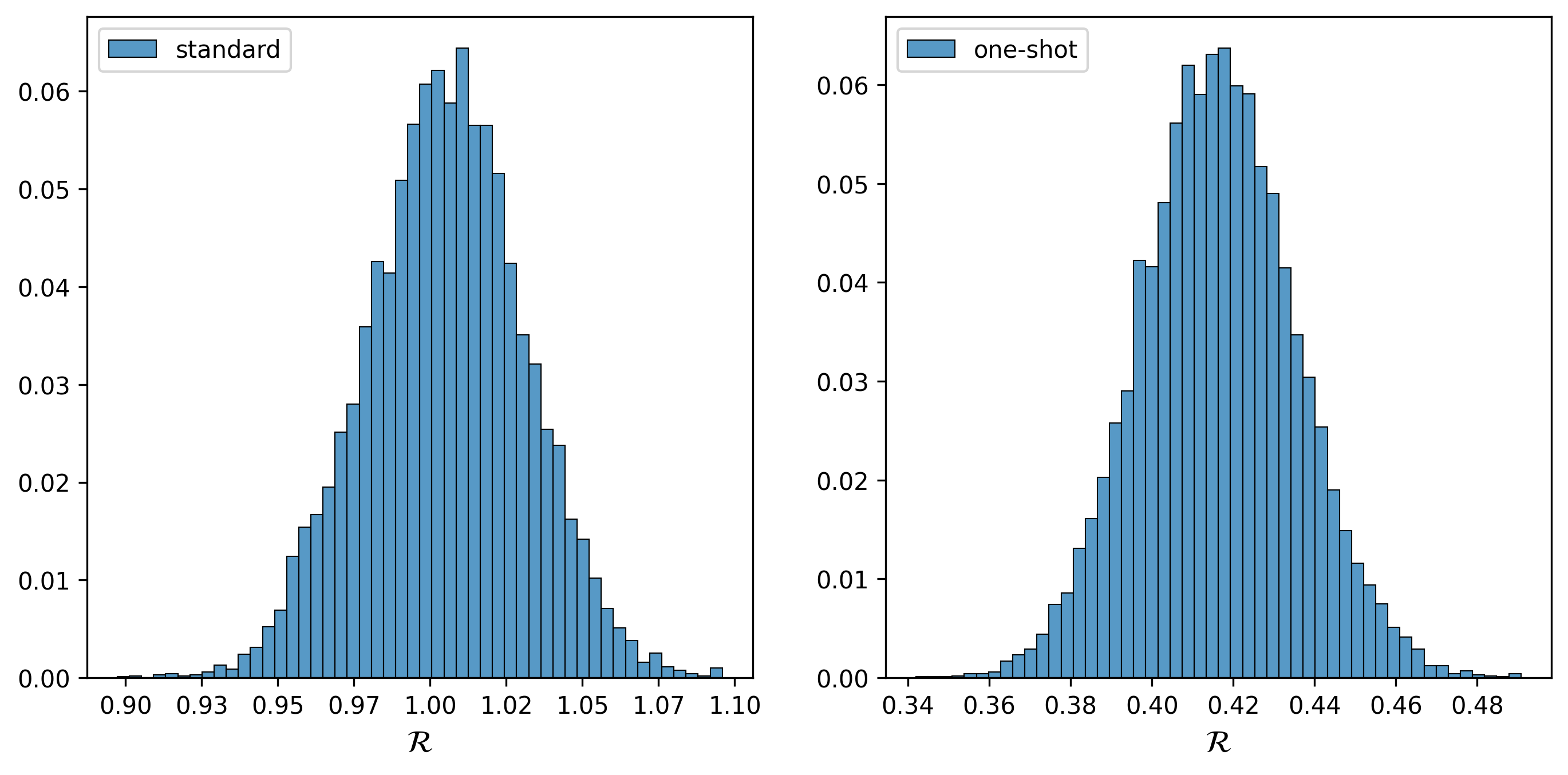}
\caption{\textcolor{blue-}{Empirical histogram of the effective range $\mathcal{R}$ of random feature maps when only good rows are used with $p_g=1$ for the standard hit-and-run Algorithm~\ref{algo:hr} (left) and the one-shot hit-and-run Algorithm~\ref{algo:hr-D} (right). Each histogram represents $10,000$ samples differing in the internal weights, the training data and the validation data. The feature dimension is $D_r=512$ and a regularization parameter of $\beta = 2.79\times10^{-5}$ is used with training data length $N=20,000$. For the standard hit-and-run Algorithm~\ref{algo:hr}, $10$ decorrelation steps are used. For the standard hit-and-run algorithm we obtain a mean forecast time of $\mathbb{E}[\tau_f] = 1.0$ with $\sigma[\tau_f]=0.03$. For the one-shot algorithm we obtain $\mathbb{E}[\tau_f] = 0.42$ with $\sigma[\tau_f]=0.02$.}}
    \label{fig:R-dist}
\end{figure}

The hit-and-run Algorithms~\ref{algo:hr} and \ref{algo:hr-D} were designed to uniformly sample from the set \textcolor{blue-}{$\Omega_{g}$}. This does not imply, however,  that ${\bb{w}}_{\rm{in}} {\bb{u}}+{\bb{b}}_{\rm{in}}$ is uniformly distributed in the interval $(-L_1,-L_0) \cup (L_0,L_1)$. To quantify the occupied range of random features we introduce the following notation. A sample $(\mathbf{w}_i^{\rm in},b_i^{\rm in})$ produces outputs the absolute values of which lie in the interval $[m_i, M_i]$, i.e.
\begin{equation}
\begin{aligned}
    m_i=\min_{1\le n\le N}|\bb{w}_{i}^{\rm in}\cdot\mathbf{u}_n+b_i^{\rm in}|,\\
    M_i=\max_{1\le n\le N}|\bb{w}_{i}^{\rm in}\cdot\mathbf{u}_n+b_i^{\rm in}|.
\end{aligned}
\label{eq:eff-lim}
\end{equation}
The effective range $\mathcal{R}$ of a random feature map vector can then be defined as
\begin{align}
    \mathcal{R} = \frac{1}{D_r}\sum_{i=1}^{D_r}(M_i-m_i).
    \label{eq:Rr}                             
\end{align}
By the central limit theorem, for $D_r\gg 1$ the effective range \eqref{eq:Rr} approaches a normally distributed random variable. Figure~\ref{fig:R-dist} shows \textcolor{blue-}{the histogram of the range $\mathcal{R}$ for the standard and the one-shot hit-and-run algorithms when only good features are used with $p_g=1$. As expected both algorithms generate near-Gaussian distributions of $\mathcal{R}$.}

\begin{figure}[!htp]
    \centering
    \includegraphics[scale=0.8]{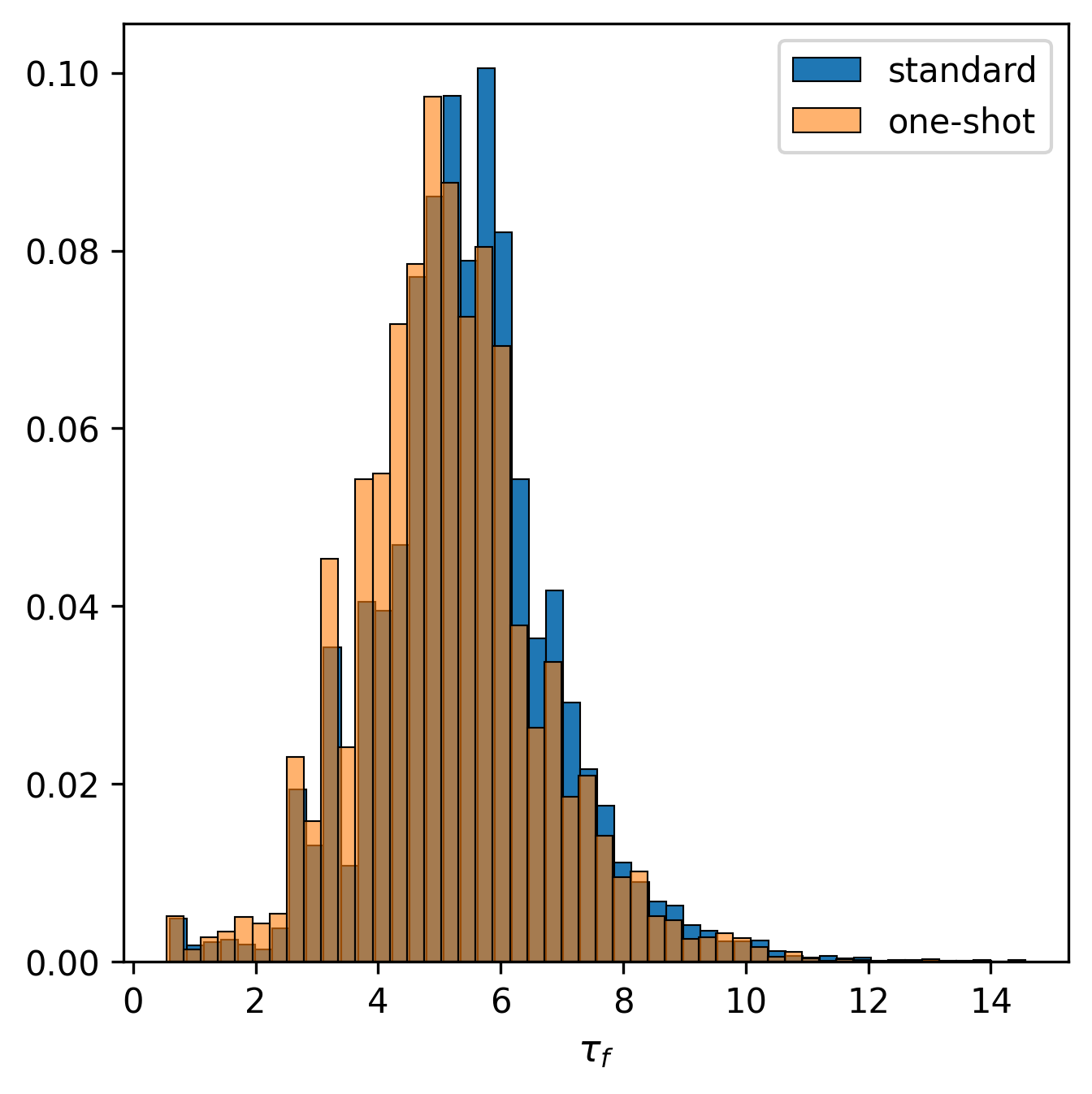}
    \caption{\textcolor{red-}{Empirical histogram of the forecast time $\tau_f$ when only good rows are used with $p_g=1$ for the standard hit-and-run Algorithm~\ref{algo:hr} and the one-shot hit-and-run Algorithm~\ref{algo:hr-D}. The same random feature maps are used as in Figure~\ref{fig:R-dist}. For the standard hit-and-run algorithm we obtain the mean forecast time $\mathbb{E}[\tau_f] = 5.4$ with $\sigma[\tau_f]=1.5$. For the one-shot algorithm we obtain $\mathbb{E}[\tau_f] = 5.1$ with $\sigma[\tau_f]=1.5$. The random feature maps shown here are the same as those in Figure~\ref{fig:R-dist}.}}
    \label{fig:eff-act-range}
\end{figure}

\textcolor{blue-}{The standard hit-and-run algorithm generates a wider range with $\mathbb{E}[\mathcal{R}]=1.0$ and standard deviation $\sigma[\mathcal{R}] = 0.03$ compared to the one-shot hit-and-run algorithm with $\mathbb{E}[\mathcal{R}]=0.42$ and standard deviation $\sigma[\mathcal{R}] = 0.02$. One would like the range $\mathcal{R}$ to be as large as possible, assuming that higher variability in the features increases the expressivity of the random feature map, and hence the forecast skill. Figure~\ref{fig:eff-act-range} shows the histogram of the forecast times $\tau_f$ for the two algorithms. The standard hit-and-run algorithm exhibits a slightly better forecast skill as measured by an approximately $6\%$ larger mean forecast time of $\mathbb{E}[\tau_f]=5.4$ compared to that of the one-shot hit-and-run algorithm with $\mathbb{E}[\tau_f]=5.1$, in accordance with our intuition that larger range is beneficial.}\\

\textcolor{blue-}{The one-shot hit-and-run Algorithm~\ref{algo:hr-D} is $\sim15$ times faster than the standard hit-and-run Algorithm~\ref{algo:hr} (with $10$ decorrelation steps) on an M1 CPU. Since the one-shot algorithm does not involve loops for each realization, it allows for efficient parallelization on GPUs. The standard hit-and-run algorithm, on the other hand, involves loops to execute the bisection, which prohibits effective parallelization on GPUs. As a result, the one-shot algorithm is $\sim825$ faster than the standard algorithm on an A100 GPU. In applications we hence use the computationally more efficient one-shot hit-and-run Algorithm~\ref{algo:hr-D}, given the almost equal performance in forecasting (cf.Figure~\ref{fig:eff-act-range}).}

%


\section{Results for forecasting individual trajectories}
\label{sec:res}
In this section we explore how increasing the number of good features improves the forecasting skill of a surrogate map for the Lorenz-63 system (\ref{eq:L63}), and conversely explore the effect of linear and saturated features. To do so we define the number of good, linear and saturated features in a random feature vector of dimension $D_r$ as 
\begin{align}
N_g = p_g D_r, \qquad N_l = p_l D_r, \qquad N_s = p_s D_r,
\end{align}
where the respective fractions satisfy $p_g+p_l+p_s = 1$. We construct random feature maps with internal weights $(\bb{W_{\rm in}},\bb{b}_{\rm in})$ with specified fractions of good, linear or saturated rows using the one-shot hit-and-run Algorithm~\ref{algo:hr-D}. 


\subsection{Effect of the quality of internal weights on the forecast time \texorpdfstring{$\tau_f$}{Lg}}
\label{ssec:tau_f-vs-p_g} 
In this section we investigate how the forecasting skill of a random feature surrogate model (\ref{eq:dsc-surrogate}) improves with increasing number of good rows $N_g$. We would like to have internal parameters resulting in large mean forecast times $\tau_f$ with relatively small standard deviations. For chaotic dynamical systems we expect a residual variance of the forecast time due to the sensitivity to small changes in the model: small changes in the internal parameters may cause the surrogate models to deviate from each other after some time. 

We estimate the mean of the forecast time $\tau_f$ and its coefficient of variation as a function of the fraction of good features $p_g$, varying $p_g$ from $p_g=0$ with only bad features to $p_g=1$ with only good features present. For each value of $p_g$ we approximately uniformly distribute the remaining $(1-p_g)D_r$ features over the linear and saturated features with $p_l\approx p_s\approx (1-p_g)/2$. Note that we cannot always impose perfect equality since $N_g=p_gD_r$, $N_l = p_l D_r$ and $N_s = p_s D_r$ are integers. We use $51$ equally spaced values of $p_g$ in $[0, 1]$ and compute averages over $500$ realizations for each value of $p_g$, differing in the draws of the random internal weights, the training data and the validation data.

\begin{figure}[!htp]
    \centering
\includegraphics[scale=0.6]{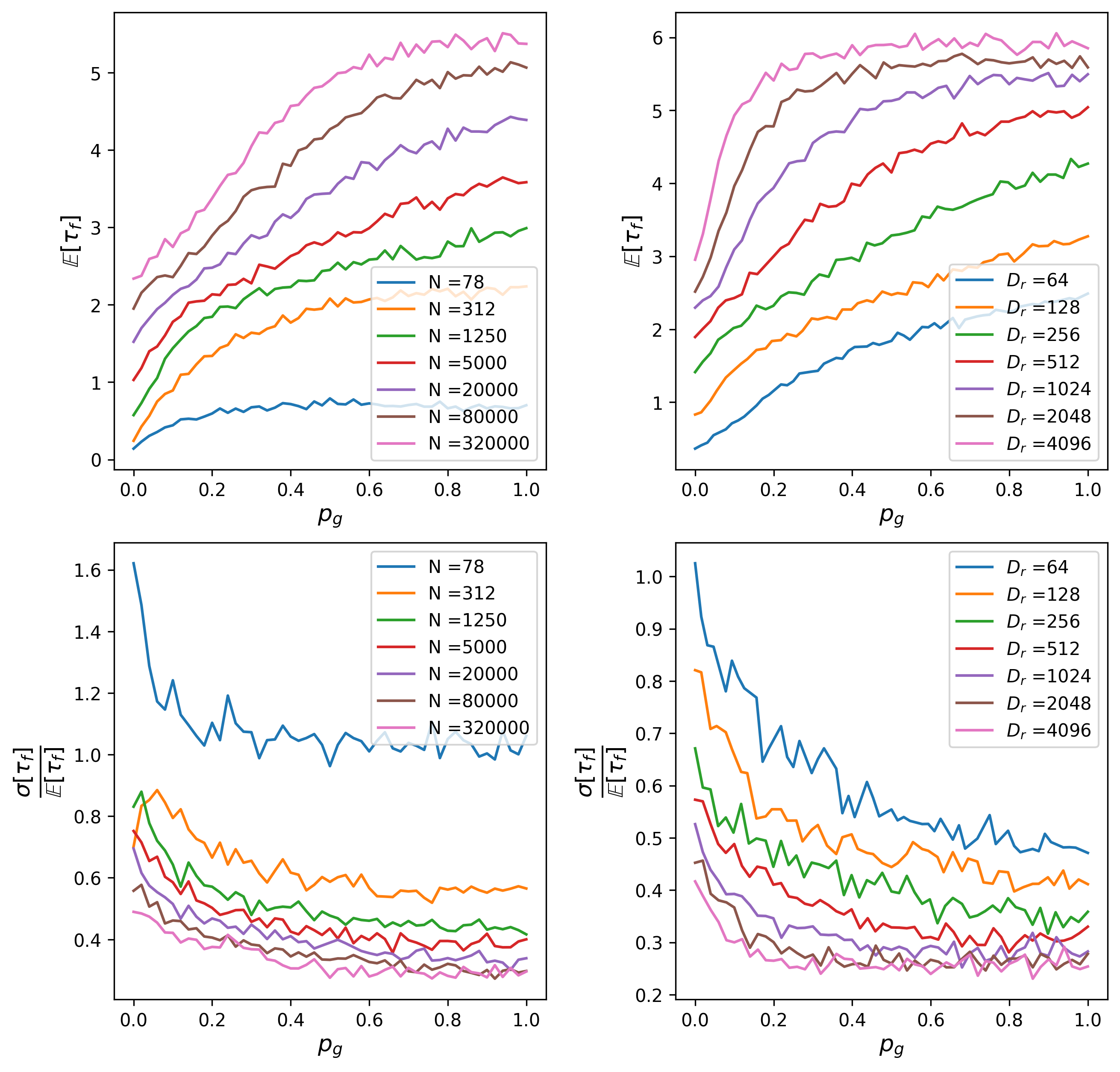}
    \caption{The top row depicts the mean of the forecast time $\mathbb{E}[\tau_f]$ as a function of the fraction of good features $p_g$. The bottom row depicts the coefficient of variation $\sigma[\tau_f]/\mathbb{E}[\tau_f]$ as a function of $p_g$. Along the first column the feature dimension $D_r=300$ is kept constant, and along the second column the length of the training data set $N=20,000$ is kept constant. Expectation are computed over $500$ realizations of the internal parameters, the training data and testing data. A regularization parameter of $\beta=4\times 10^{-5}$ is employed.}
    \label{fig:tau_f-vs-p_g}
\end{figure}

\begin{figure}[!htp]
    \centering
    \includegraphics[scale=0.6]{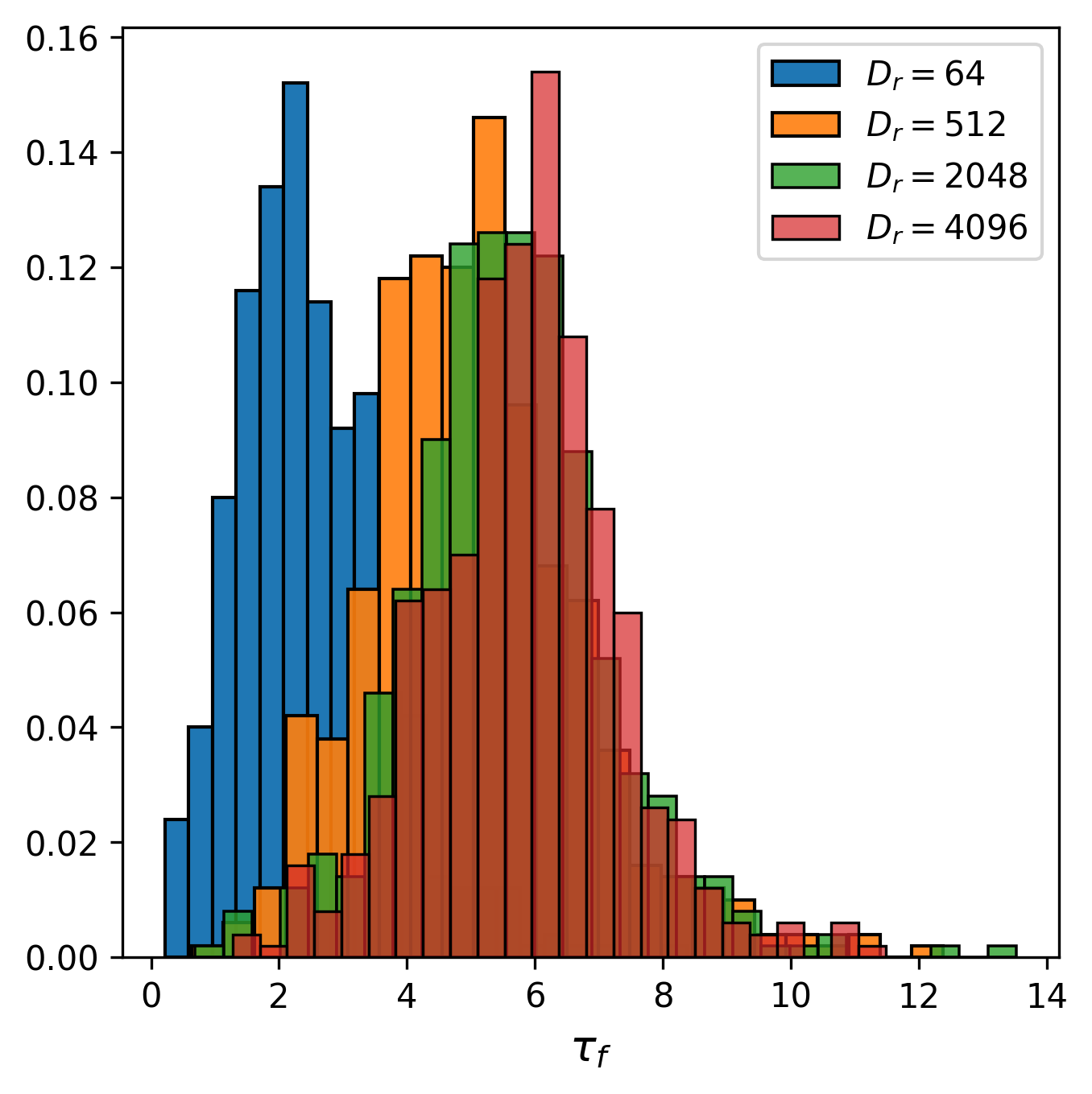}
    \caption{Empirical histogram of $\tau_f$ for different values of $D_r$ when $p_g=1$ for increasing feature dimension $D_r$. The same $500$ realizations are used as in Figure~\ref{fig:tau_f-vs-p_g} with $N=20,000$.}
    z
    \label{fig:tau_f-dist-evol}
\end{figure}

Figure~\ref{fig:tau_f-vs-p_g} shows the dependence of the mean forecast time $\mathbb{E}[\tau_f]$ and the associated coefficient of variation $\sigma[\tau_f]/\mathbb{E}[\tau_f]$ on $p_g$ for various values of the feature dimensions $D_r$ and training data lengths $N$. It is clearly seen that increasing the number of good rows increases the mean forecast time and decreases the coefficient of variation, as desired. As expected, for fixed feature dimension $D_r$ increasing the training data length $N$ is beneficial. \textcolor{red-}{A training data length of $N=78$ is too small to provide any meaningful forecasting skill with mean forecast times below one Lyapunov time for any value of $p_g$ and significantly larger variance.} Similarly, for fixed training data length $N$, increasing the feature dimension $D_r$ is beneficial. The observation that, for fixed data length $N$, the mean forecast time $\mathbb{E}[\tau_f]$ saturates upon increasing $p_g$ once a sufficiently large number of good features $N_g=p_g D_r$ are present, suggests that the distribution of the forecast time $\tau_f$ converges reflecting a residual uncertainty of the chaotic surrogate model. This is confirmed in Figure~\ref{fig:tau_f-dist-evol} where we see convergence of the empirical histograms of the forecast time for increasing values of $D_r$ in the case when $p_g=1$. 

\begin{figure}[!htp]
    \centering
    \includegraphics[scale=0.6]{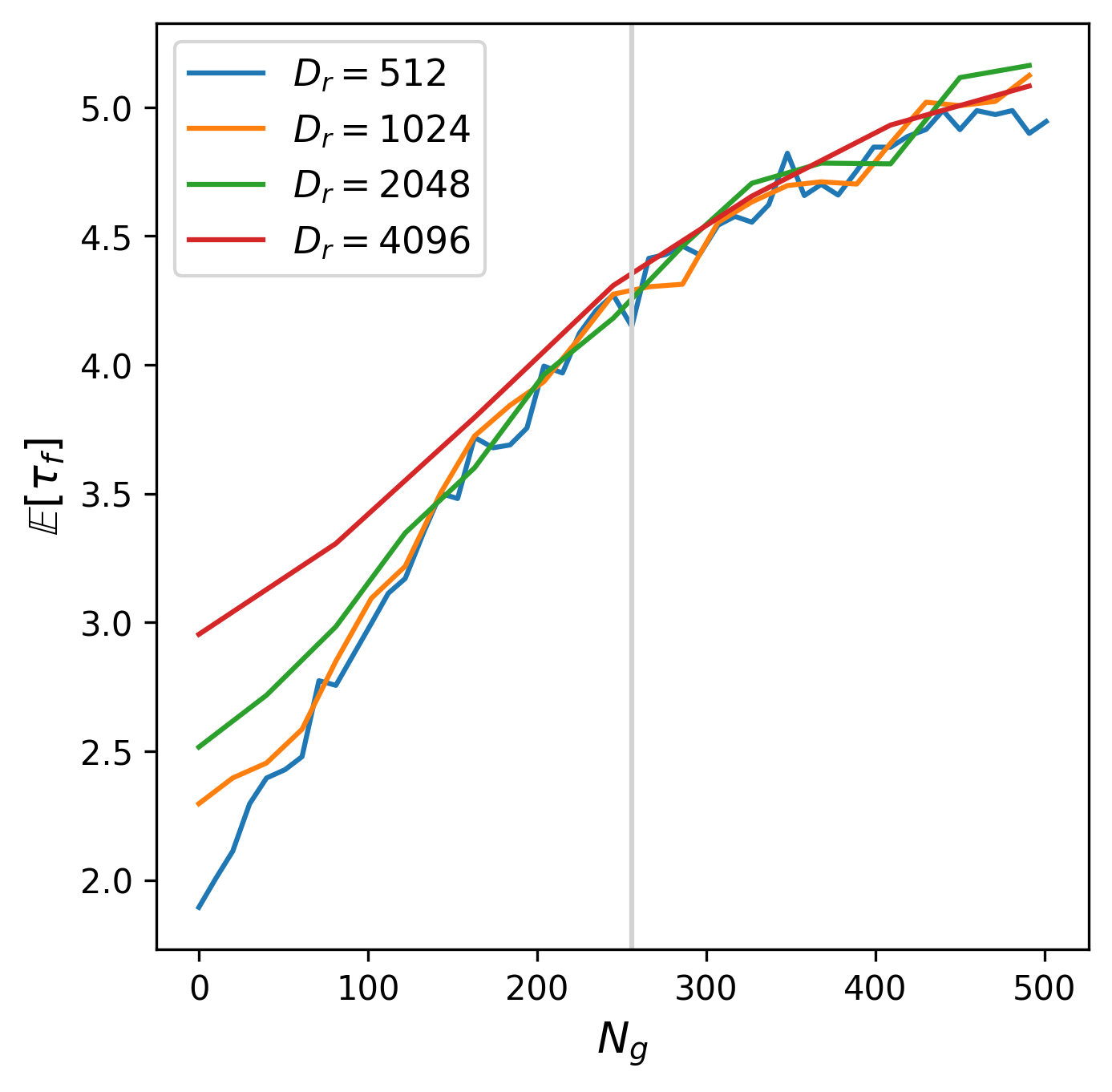}
    \caption{Forecast time mean $\mathbb{E}[\tau_f]$ as a function of good features $N_g=p_gD_r$. The vertical line demarcates $N_g=256$. The range of $N_g$ is restricted to $N_g\le 512$, corresponding to $p_g=1$ for the smallest value of the feature dimension $D_r=512$. The same $500$ realizations are used as in Figure~\ref{fig:tau_f-vs-p_g} with $N=20,000$.}
    \label{fig:effective-D_r}
\end{figure}

\begin{figure}[!htp]
    \centering
    \includegraphics[scale=0.52]{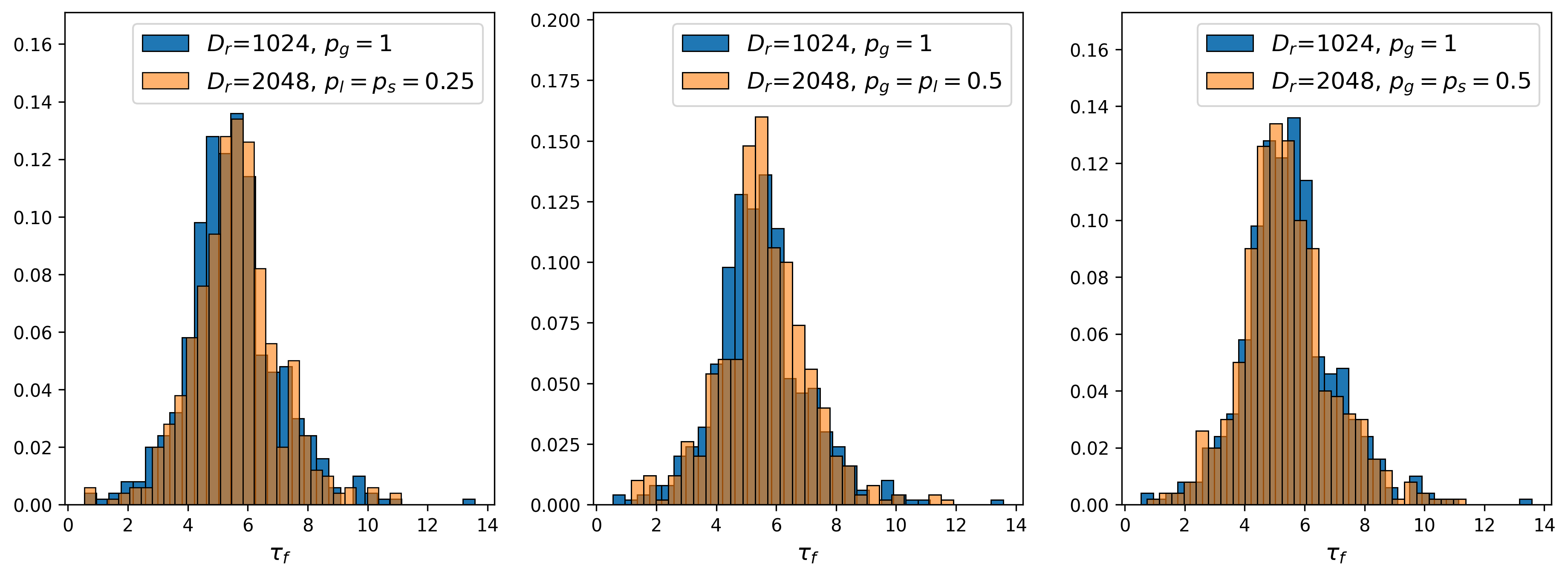}
    \caption{Empirical histogram of the forecast time $\tau_f$ for $D_r=1,024$ and $D_r=2,048$. In each case the number of good rows is $N_g=1,024$. For $D_r=2,048$ we show results for an equal number of linear and saturated features with $p_l=p_s=0.25$ (left), for only linear bad features with $p_l=0.5,p_s=0$ (middle) and for only saturated bad features with $p_s=0.5,p_l=0$ (right) for $D_r=2,048$. We used $500$ realizations differing in the random draws of the internal parameters, the training data and the validation data. We employed a regularization parameter of $\beta=4\times10^{-5}$ and used training data of length $N=20,000$.}
    \label{fig:tau_f-gls}
\end{figure}

Figure~\ref{fig:effective-D_r} shows the dependency of the mean forecast time $\mathbb{E}[\tau_f]$ on the fraction of good rows $p_g$ for different values of $D_r$. We can clearly see that beyond $N_g=p_gD_r=256$ (indicated by the vertical line), the mean forecast time $\mathbb{E}[\tau_f]$ depends only on the number of good rows $N_g=p_gD_r$ and not on the overall feature dimension $D_r$. For smaller number of good rows $N_g<256$ the mean forecast time depends on the feature dimension $D_r$ with larger feature dimensions implying larger mean forecast times. This suggests that the number of good features $N_g$ constitutes an effective feature dimension $D_r^*$, which controls the forecast skill of the learned surrogate model. This implies that on average the forecast time $\tau_f$ is the same for a random feature surrogate model of dimension $D_r$ with only good features $p_g=1$ as a surrogate map with a larger feature dimension $\alpha D_r$ with $\alpha>1$ but only a fraction of $1/\alpha$ good rows. This is confirmed in Figure~\ref{fig:tau_f-gls} which shows the empirical histogram of $\tau_f$ for fixed number of good features $N_g=p_g D_r=1,024$. We compare the distribution of the forecast times for random feature maps with $D_r=1,024$ and $p_g=1$ to those with $D_r=2,048$ and $p_g=0.5$. We show examples when the remaining bad features are either equally distributed between linear and saturated features, or only linear or only saturated. The distributions for all three examples are very similar and match the one with the smaller feature dimension but same number of good features. This leads us to conclude that the number of good rows is the only determining factor for the distribution of $\tau_f$ (all other parameters being equal), and that linear and saturated rows are equally ineffective in terms of the forecasting skill. 


\begin{figure}[!htp]
    \centering
    \includegraphics[scale=0.6]{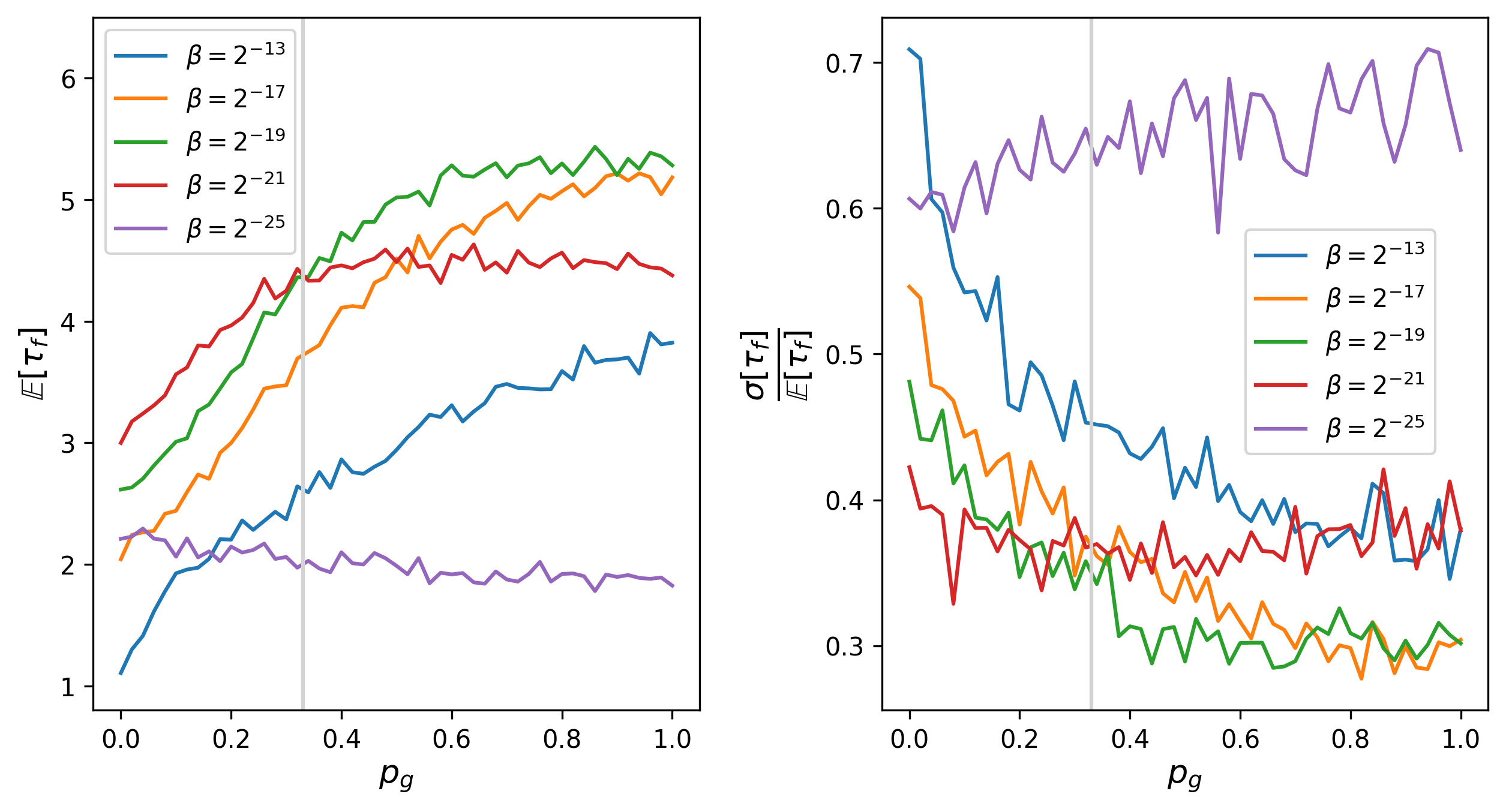}
    \caption{Mean forecast time $\mathbb{E}[\tau_f]$ (left) and coefficient of variation $\sigma[\tau_f]/\mathbb{E}[\tau_f]$ (right) as a function of $p_g$ for a range of regularization parameters $\beta$. A regularization parameter of $\beta=2^{-19}$ is optimal among the values presented here for $p_g>0.33$ (demarcated by a vertical line). Results are shown for fixed $D_r=300$ and $N=20,000$.}
    \label{fig:tau_f-beta}
\end{figure}

We briefly discuss the effect of the regularization parameter $\beta$ on the forecasting skill. We show in Figure~\ref{fig:tau_f-beta} the mean forecast time $\mathbb{E}[\tau_f]$ and coefficient of variation $\sigma[\tau_f]/\mathbb{E}[\tau_f]$ as a function of $p_g$ for a range of regularization parameters $\beta \in [2^{-25},2^{-13}]$. For fixed feature dimension $D_r=300$, we see that $\beta=2^{-19}$ is optimal within this range in terms of the mean forecast time (left panel) and the coefficient of variation (right panel) once sufficiently many good features are present with $p_g>0.33$. Note that we had previously employed $\beta=4\times 10^{-5}\approx 2^{-14.6}$. \textcolor{blue-}{The large difference in performance for different choices of the regularization parameter $\beta$ makes clear that, if optimizing for performance, the regularization parameter has to be optimized. This could be achieved by line-search \cite{MandalGottwald25} or by Bayesian optimization \cite{LevineStuart22}. Here we have refrained from fine-tuning the regularization parameter as our focus is the sampling algorithm and the effect of different types of features on the forecast skill.}

%


\subsection{Effect of the quality of internal weights on the outer weights \texorpdfstring{$\bb{W}$}{W}}
\label{ssec:W-vs-p_g} 

In this section we explore how the nature of the internal weights affects the learned solutions of the ridge regression (\ref{eq:sol-lr}) which we denote simply as $\bb{W}$, dropping the star. 

\begin{figure}
    \centering
    \includegraphics[scale=0.6]{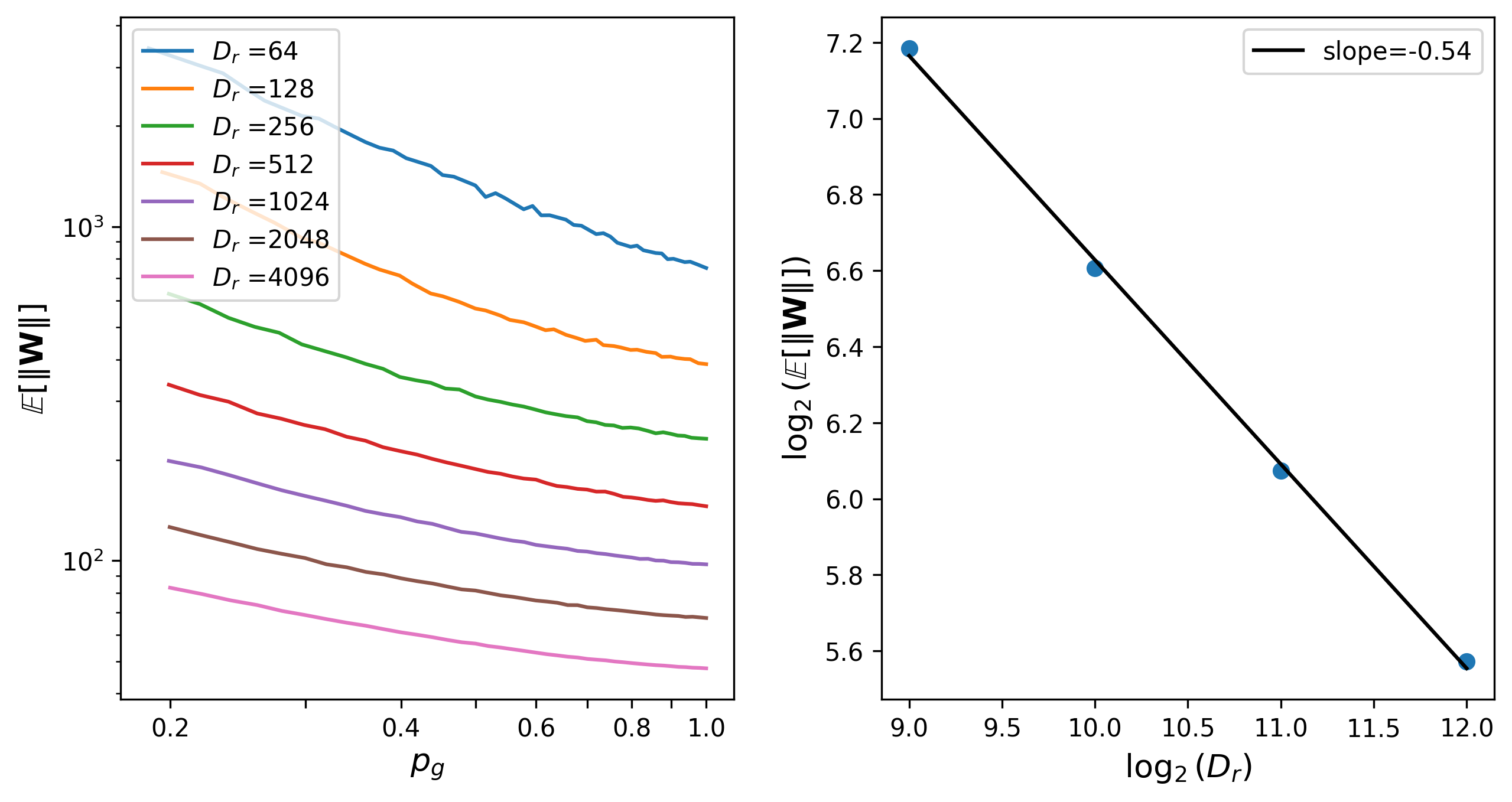}
    \caption{Left: The mean of the Frobenius norm of the outer weights, $\|\bb{W}\|$, as a function of $p_g$ on a log-log scale. Right: The mean of the Frobenius norm of the outer weights, $\|\bb{W}\|$, as a function of the feature dimension $D_r$ for $p_g=1$. The line of best fit with approximate slope $-0.54$ is also shown on the right. 
    The data are from the same experiments as shown in Figure~\ref{fig:tau_f-vs-p_g}.}
    \label{fig:W-vs-p_g-512}
\end{figure}

We begin by recording the Frobenius norm $\bb{\|W\|}={\rm{Tr}}({\bb{W}}^\top {\bb{W}})$ of the learned outer weights as a function of the fraction of good features $p_g$ (bad features are again roughly equally distributed between linear and saturated features). Figure~\ref{fig:W-vs-p_g-512} shows the mean of $\bb{\|W\|}$ as a function of $p_g$ on a $\log$-$\log$ scale for the simulations used in Figure~\ref{fig:tau_f-vs-p_g}. It is seen that $\|\bb{W}\|$ is a decreasing function of the number of good features. The solution of the linear regression problem $\bb{W}$ minimizes the loss function (\ref{eq:cost}). Once there are sufficiently many good features, the training data can be sufficiently well fit, decreasing the first term of the loss function. Further increasing the number of good features then allows for a decrease of the regularizing term of the loss function, leading to a decrease of $\|\bb{W}\|$. Assuming that the true one-step map $\Psi_{\Delta t}$ in (\ref{eq:dsc-system}) lies in the domain of the random feature map (\ref{eq:dsc-surrogate}) with infinitely many features, the first term of the loss function should scale with the usual Monte-Carlo estimate scaling of $\mathcal{O}(1/D_r)$, suggesting a scaling of the regularization term $\|\bb{W}\|\sim 1/\sqrt{D_r}$. In the right panel of Figure~\ref{fig:W-vs-p_g-512} we show that the mean of $\|\bb{W}\|$ roughly scales as $\|\bb{W}\|\sim 1/D_r^{0.54}$ when all the internal weights correspond to good features with $p_g=1$, suggesting that the true one-step map can be well approximated by random features with a $\tanh$-activation function. We remark that the Monte-Carlo scaling is valid for $D_r>256$ only, i.e. provided sufficiently many good features are present.\\

We now investigate how the decrease in the outer weights $\bb{W}$ is distributed over the various features. We will see that the outer weights are learned to suppress the bad features provided there are sufficiently many good features allowing for a reduction of the loss function. Let us denote the $i$-th column of $\bb{W}$ by $\bb{W}^i$. The columns $\bb{W}^i$ are the weights attributed to the features produced by the $i$-th row of the internal weights $(\bb{w}^{\rm in}_i, b^{\rm in}_i)$. We expect the outer weights corresponding to good rows to be significantly larger than those corresponding to bad rows.

To study the suppression of bad features by columns of $\bb{W}$ which have small norm, we design two sets of numerical experiments: one in which bad features are entirely comprised of linear features and one in which bad features are entirely comprised of saturated features. 

\begin{figure}[!htp]
    \centering
    \includegraphics[scale=0.61]{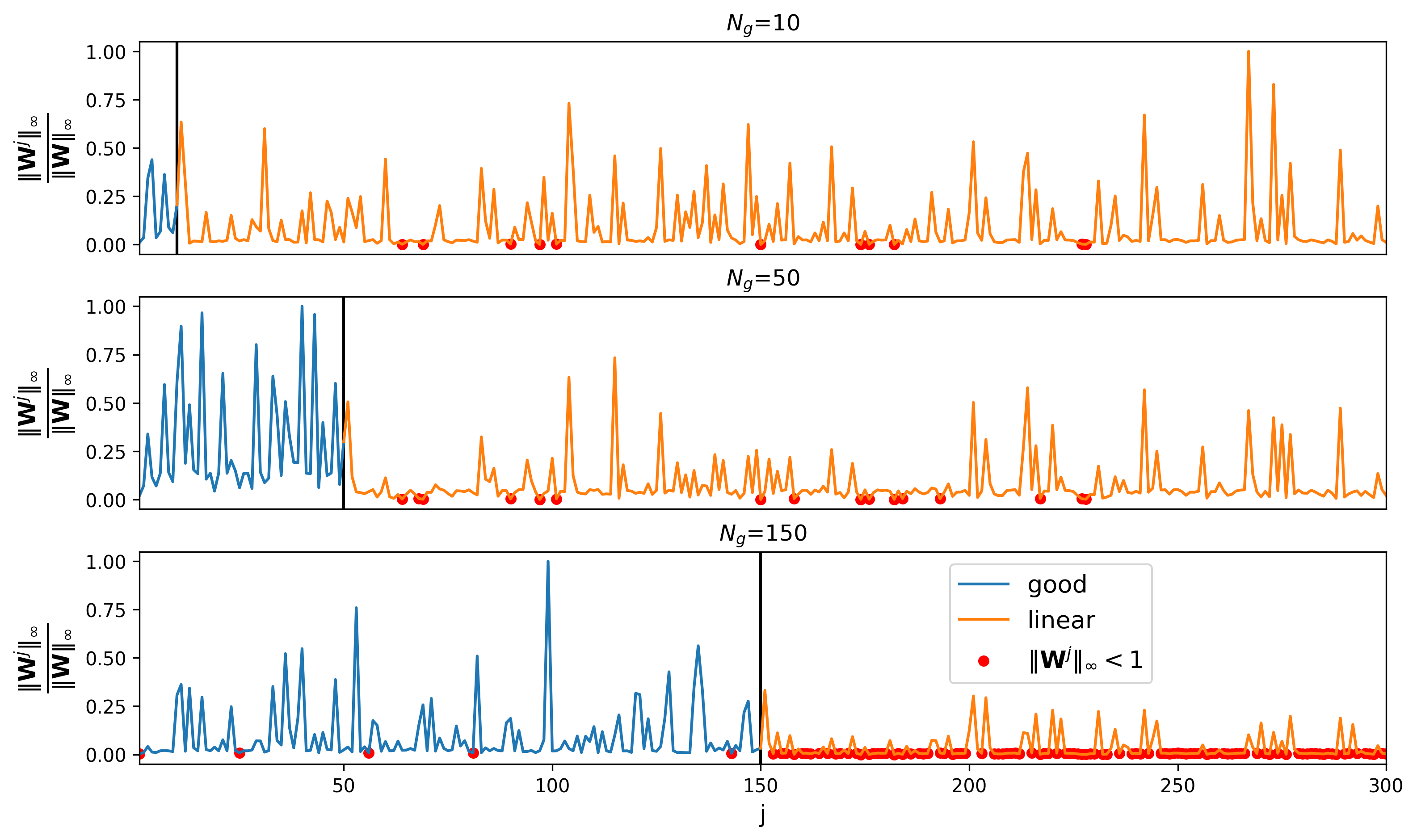}
    \caption{Normalized supremum norm of the columns of $\bb{W}$ for different numbers of good features with $N_g=10$, $N_g=50$ and $N_g=150$ and otherwise exclusively linear bad features. The $x$-axis represents column indices. The good and linear columns are indicated in blue and orange, respectively. The red dots signify columns with supremum norm less than $1$. The overall feature dimension is $D_r=300$ and the outer weights were obtained from training data of length $N=20,000$.}
    \label{fig:slider-linear}
\end{figure}

In the first set we initialize a random feature map with $D_r=300$ features consisting of only bad linear features. We then successively replace one linear feature by a good feature, i.e. replacing one inner linear weight row $(\mathbf{w}_i^{\rm in},b_i^{\rm in})$ by a good row. At each step we record the corresponding linear regression solution $\bb{W}$. Figure~\ref{fig:slider-linear} shows the normalized supremum norm of columns of $\bb{W}$ after $N_g=10$, $N_g=50$ and $N_g=150$ bad linear features have been replaced by good features. The red dots signify columns which do not contain any entry with absolute value larger than $1$. It is clearly seen that linear features are suppressed by the columns of $\bb{W}$. Note that not all linear features are entirely suppressed.

\begin{figure}[!htp]
    \centering
    \includegraphics[scale=0.61]{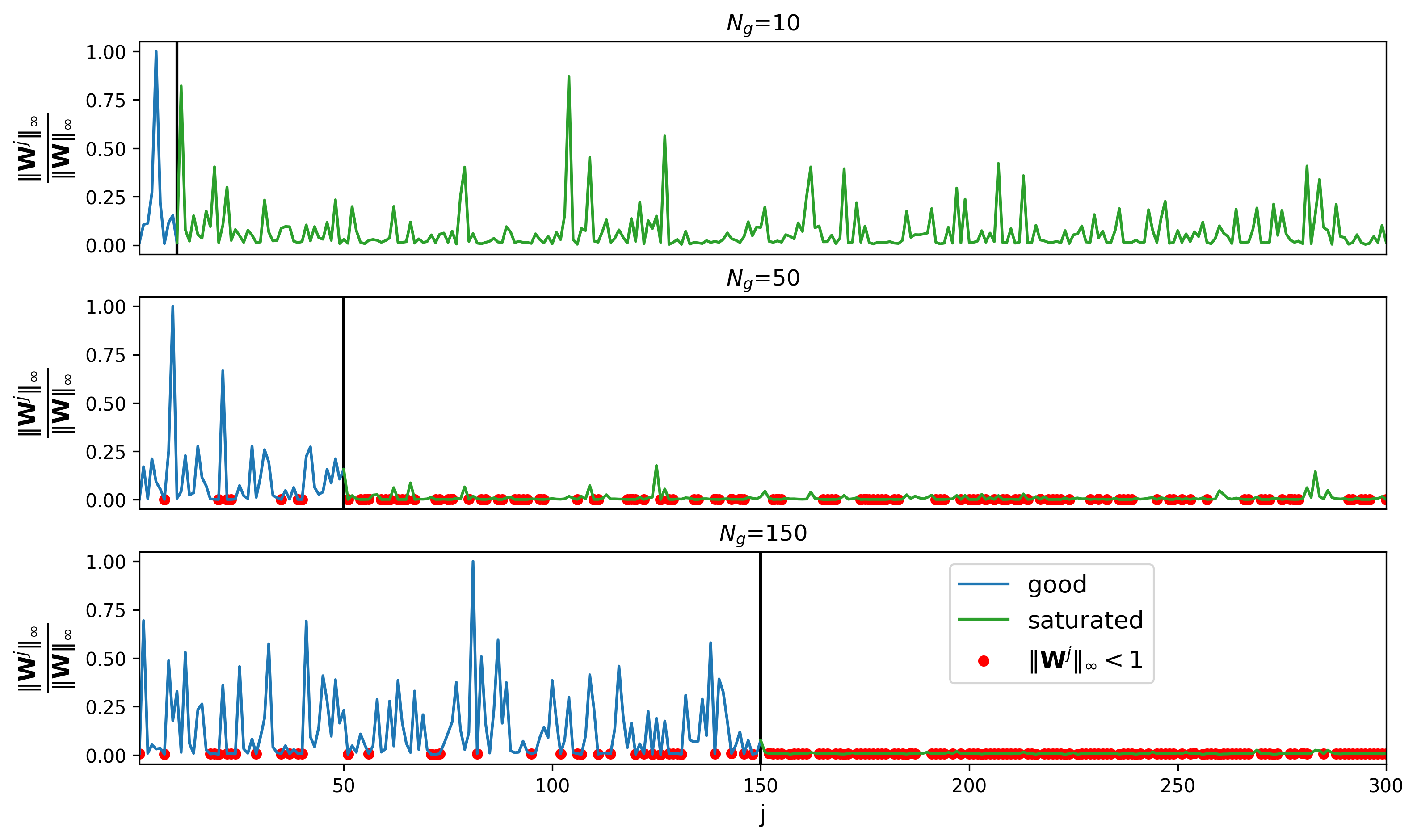}
    \caption{Normalized supremum norm of the columns of $\bb{W}$ for different numbers of good features with $N_g=10$, $N_g=50$ and $N_g=150$ and otherwise exclusively saturated bad features.  The $x$-axis represents column indices. The good and saturated columns are indicated in blue and green, respectively. The red dots signify columns with supremum norm less than $1$. The overall feature dimension is $D_r=300$ and the outer weights were obtained from training data of length $N=20,000$.}
    \label{fig:slider-saturated}
\end{figure}

In the second experiment, we follow the same procedure as before except we start with only saturated random features. In Figure~\ref{fig:slider-saturated} it is seen that saturated features are suppressed even stronger by the outer weights than linear features. In contrast to linear features, saturated features are effectively fully suppressed once the number of good features exceeds $N_g=50$.

%


\section{Comparison with a single-layer feedforward network trained with gradient descent}
\label{sec:NN}
A natural question is if a single-layer feedforward network of the architecture (\ref{eq:architecture}) for which the internal weights $(\mathbf{W}_{\rm in}, \mathbf{b}_{\rm in})$ are learned together with the outer weights $\mathbf{W}$ performs better or worse than random feature maps with fixed good internal weights. In particular, we consider the non-convex optimization problem 
\begin{align}
\mathbf{\textcolor{blue-}{\Theta^*}}= \underset{\Theta}{\argmin}\;\mathcal{L}(\textcolor{blue-}{\Theta};\bb{U}),
\label{eq:NNop}
\end{align}
with $\bb{\Theta}=(\bb{W}_{\rm in},\bb{b}_{\rm in}, \bb{W})$ and the loss function $\mathcal{L}$  defined in (\ref{eq:cost}). To solve the optimization problem (\ref{eq:NNop}) we employ gradient descent. To fairly compare with the results from the random feature model, we set the internal layer width to $D_r = 300$, the regularization parameter to $\beta = 4 \times 10^{-5}$, and the training data length to $N = 20,000$. To initialize the network weights we use the standard Glorot initialization \cite{glorot2010understanding}. We use an adaptive learning rate scheduler which is described in Appendix~\ref{sec:rate}. 

\begin{figure}[!htp]
    \centering
    \includegraphics[scale=0.6]{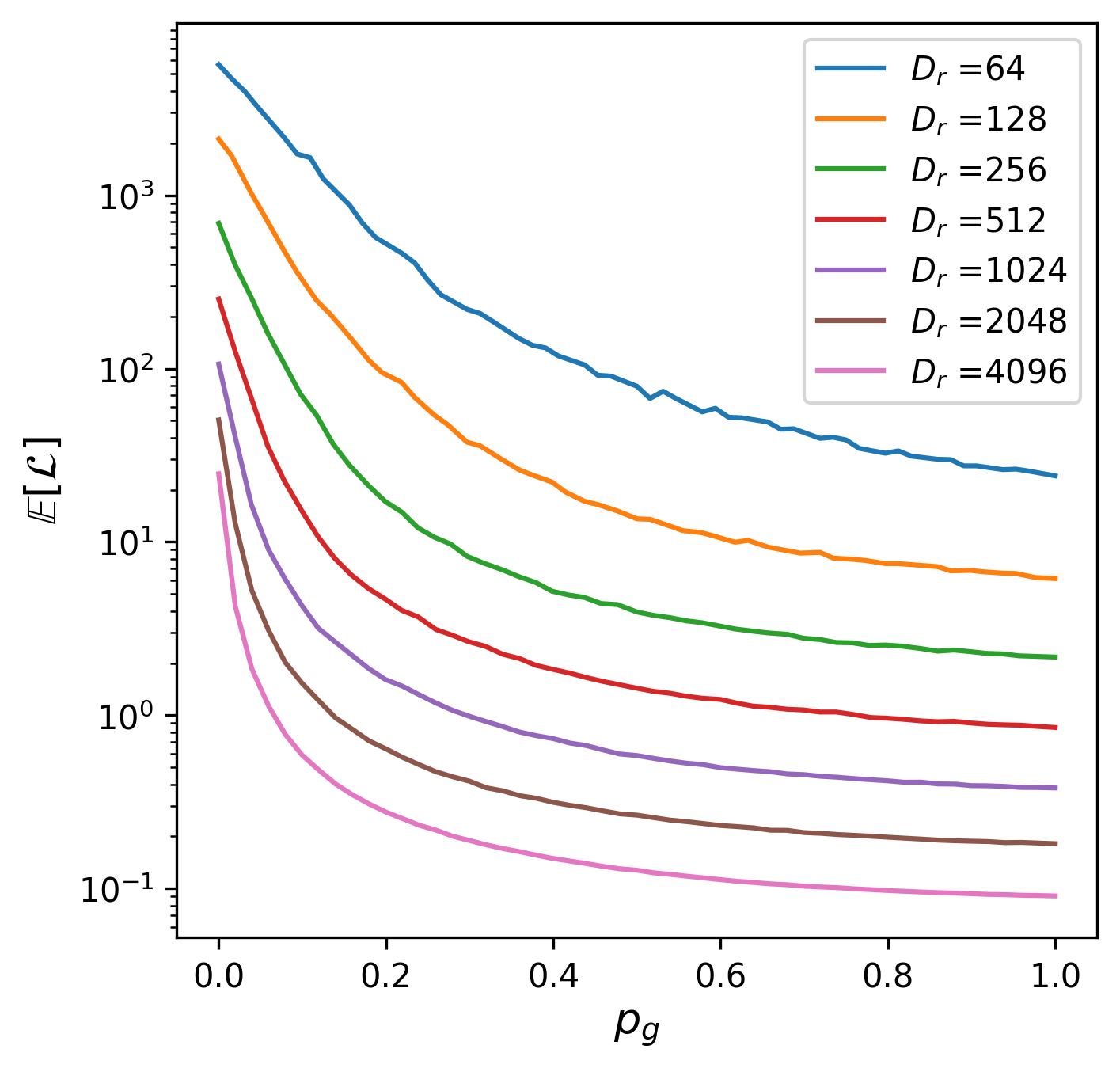}
    \caption{Mean loss ${\mathcal{L}}$ for random feature maps as a function of $p_g$ for different values of the feature dimension $D_r$. The data shown here correspond to the experiments shown in Figure~\ref{fig:tau_f-vs-p_g} with $N=20,000$ and $\beta=4\times10^{-5}.$}
    \label{fig:loss_vs_p_g}
\end{figure}

Figure~\ref{fig:nn-evol} shows the evolution of the mean forecast time $\mathbb{E}[\tau_f]$ and the logarithm of the loss function $\mathcal{L}$ during training. The expectation is computed over $500$ different validation data sets.  Optimization over all weights clearly allows for a significantly smaller training loss $\mathcal{L}$ compared to random feature maps (cf. Figure~\ref{fig:loss_vs_p_g}). The neural network achieves a final value of the loss function of $\mathcal{L} \approx 0.09$ which is a $95\%$ improvement when compared to a random feature map of the same size with only good internal parameters, i.e. $p_g=1$, which has a loss of $\mathcal{L} \approx 1.73$ on average. However, the situation is very different for the mean forecast time. The mean forecast time $\mathbb{E}[\tau_f]$ is a slowly growing function of the gradient descent steps with the last $10^5$ steps resulting in only about $0.32\%$ improvement. The data are plotted every $10^4$ steps and therefore the typical fluctuations of gradient descent are not visible.  Maybe surprisingly, optimizing the internal weights via gradient descent does not lead to a better forecasting skill when compared to the random feature map surrogate model. After $1.5\times10^6$ steps the neural network achieves a mean forecast time of only $\mathbb{E}[\tau_f]\approx3.75$. Random feature maps of the same size with $p_g=1$ generate a mean forecast time of  $\mathbb{E}[\tau_f]\approx4.46$ (cf. Figure~\ref{fig:tau_f-vs-p_g}). Furthermore, the training took approximately $8.2\times10^4$ seconds on the T4 GPU available through Google Colab cloud platform. In contrast, initializing and training a random feature map of the same size took less than $1$ second in total, i.e almost $100,000$ times faster.
\begin{figure}[!htp]
    \centering
    \includegraphics[scale=0.6]{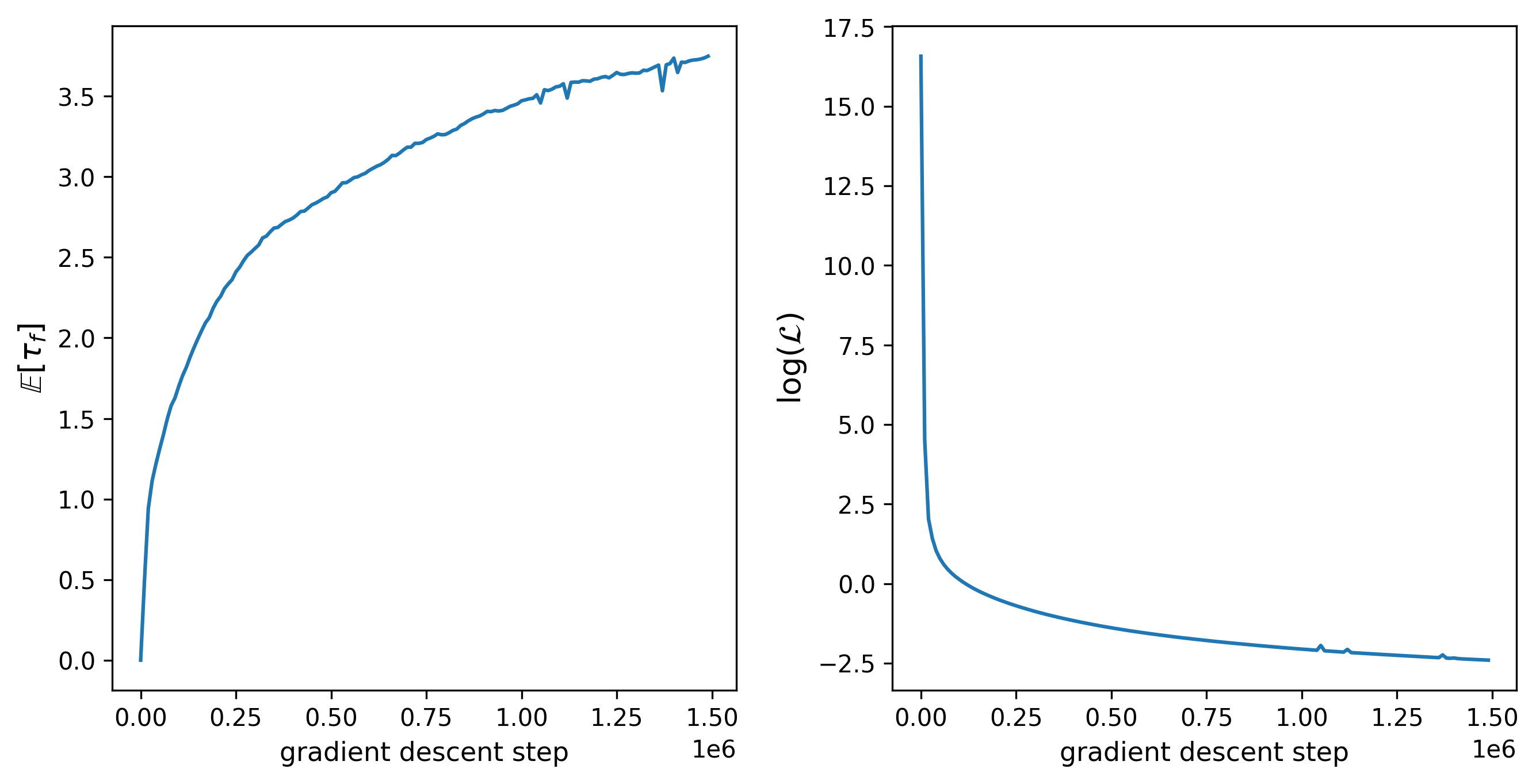}
    \caption{Evolution of the mean forecast time $\mathbb{E}[\tau_f]$ and the logarithm of the loss function (\ref{eq:cost}) $\log(\mathcal{L})$ during training of a single-layer feedforward network with gradient descent. For each step $\mathbb{E}[\tau_f]$ is computed using $500$ test trajectories. The network with width $D_r=300$ was trained with training data of length $N=20,000$ and a regularization parameter $\beta=4\times10^{-5}$. Results are shown every $10^4$ gradient descent steps.}
    \label{fig:nn-evol}
\end{figure}

\begin{figure}[!htp]
    \centering
    \includegraphics[scale=0.6]{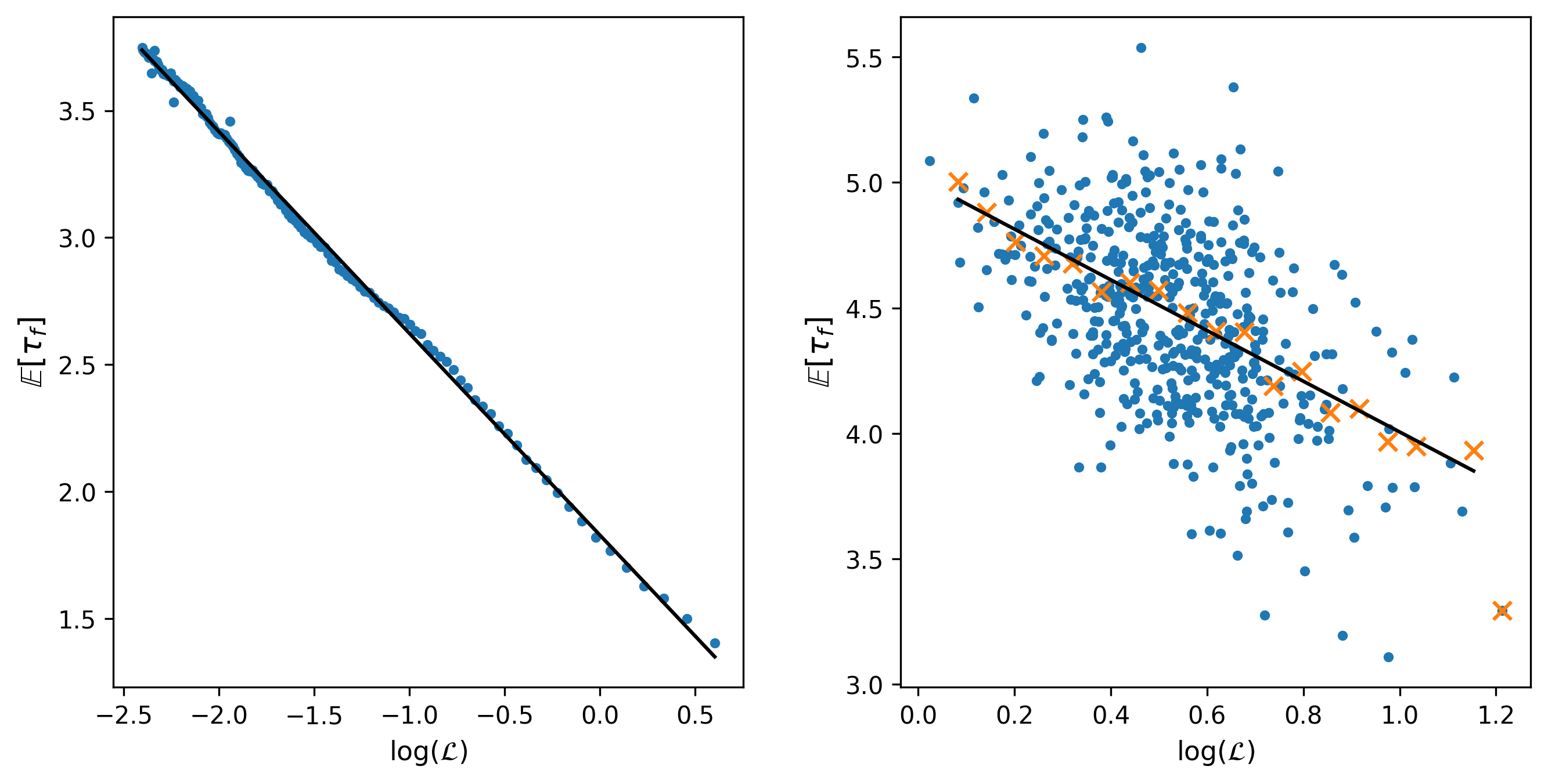}
    \caption{Relationship between the mean forecast time $\mathbb{E}[\tau_f]$ and the logarithm of the loss function $\mathcal{L}$ for a single-layer feedforward network (left) and for a random feature map with only good internal parameters, i.e. $p_g=1$ (right). Each dot in the left panel corresponds to a gradient descent step. Each dot in the right panel corresponds to one realization of a random feature map. The expectation is computed over $500$ validation trajectories. Each descent step and each realization use the same training and testing data. The black line in the left panel represents the best-fit line. 
    In the right panel the orange crosses denote the conditional mean $\mathbb E[\tau_f|\log(\mathcal L)]$ and the black line represents the best-fit line. 
    We use a feature dimension of  $D_r=300$, training data length $N=20,000$ and regularization parameter $\beta=4\times10^{-5}$.}
    \label{fig:scatter-nn-rf}
\end{figure}

The left panel of Figure~\ref{fig:scatter-nn-rf} shows that the mean forecast time $\mathbb{E}[\tau_f]$ and the logarithm of the loss function $\log(\mathcal{L})$ are linearly related for the single-layer neural network. This is a direct manifestation of the exponential sensitivity in chaotic dynamical systems: in each gradient descent step the loss experiences small changes leading to small changes in the learned weights and hence in the resulting surrogate model. These small changes in the chaotic surrogate model lead to an exponential divergence of nearby trajectories. This causes an exponential in time loss of predictability, characterized here by the mean of the forecast time~\eqref{eq:def-tau_f}. 
The same sensitive dependency on small changes of the surrogate model, quantified by small changes of the loss function, is also present in random feature maps. The right panel of Figure~\ref{fig:scatter-nn-rf} shows the mean forecast time $\mathbb{E}[\tau_f]$ as a function of the logarithm of the loss function for random feature maps. Each dot represents one realization of a random feature map with feature dimension $D_r=300$, trained on the same data as the single-layer feedforward network. The mean forecast time $\mathbb{E}[\tau_f]$ is computed using the same $500$ validation trajectories as the network. Averaged over bins of the logarithm of the loss function, the mean forecast time shows the same linear relationship with the logarithm of the loss function (orange crosses in Figure~\ref{fig:scatter-nn-rf}). The slopes of the best-fit lines in~Figure~\ref{fig:scatter-nn-rf} show that the forecasting skill of the random feature map improves slightly faster with decreasing loss when compared to the single-layer network \textcolor{red-}{with an estimated slope of $-1.01$ for the random feature map and $-0.79$ for the neural network.} 

\begin{figure}[!htp]
     \centering
     \includegraphics[scale=0.6]{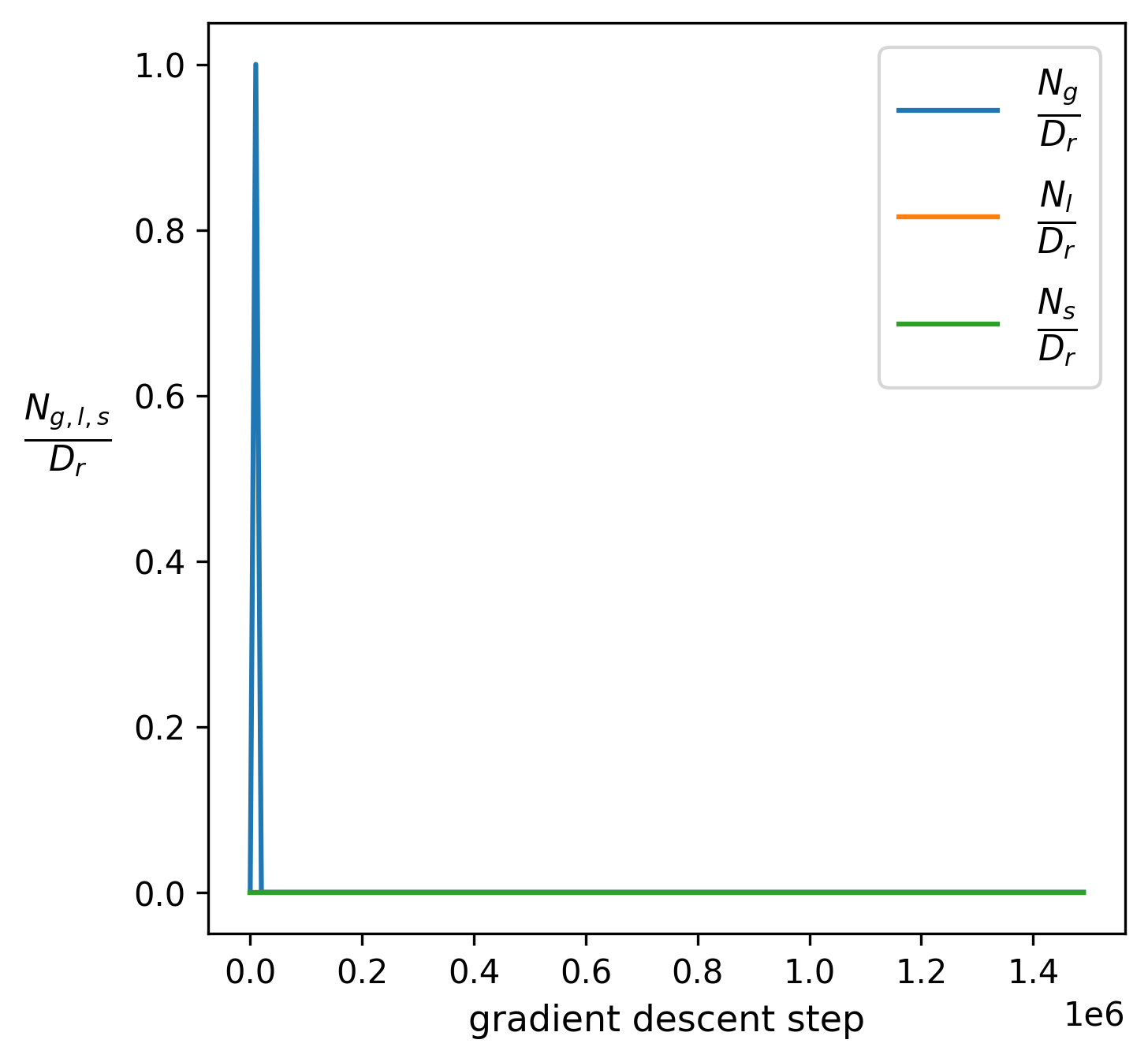}
     \caption{Evolution of the number (normalized by $D_r$) of learned good, linear and saturated rows $(\mathbf{w}_i^{\rm in},b_i^{\rm in})$ in the internal parameters during a single  training episode of a single-layer feedforward network. \textcolor{red-}{The data shown here corresponds to the training instance shown in Figure~\ref{fig:nn-evol} where we used $D_r=300$ and $N=20,000$}. Results are shown every $10^4$ gradient descent steps. 
     }
     \label{fig:no-structure}
\end{figure}

The discrepancy between the neural network having worse forecasting skill compared to random feature maps despite achieving smaller loss can be explained as follows. Minimizing the loss function $\mathcal{L}$ aims at learning the single-step surrogate map~\eqref{eq:architecture}. High forecasting skill, however, requires multiple applications of the single-step surrogate model which is not explicitly accounted for in the loss function (\ref{eq:cost}). In Section~\ref{sec:feature} we established that the main controlling factor for achieving high forecasting skill is the number of good features. In random feature maps we can control and maximize this number simply by sampling good parameters according to our hit-and-run Algorithms~\ref{algo:hr} and \ref{algo:hr-D}, respectively. On the other hand, the training of the single-layer feedforward network is only designed to minimize the loss but not to maximize the number of good features to $N_g=D_r=300$ in our case. Figure~\ref{fig:no-structure} shows the number of different types of rows produced during the training instance of Figure~\ref{fig:nn-evol}. We see that only a single good row was produced in $(\bb{W}_{\rm in},\bb{b}_{\rm in})$ during the early steps of the optimization, and this good row was then quickly destroyed during the training process. We checked that even when the network is initialized with only good internal parameters, i.e. $p_g=1$, training eventually leads to a complete absence of good internal parameters with $p_g=0$ for reasonable learning rates. To understand the absence of good rows in the trained network, note that for any random feature map  $\bb{\Theta}=(\bb{W}_{\rm in},\bb{b}_{\rm in}, \bb{W})$ essentially lies on the graph of a continuous function due to the intricate relationship between the internal and outer weights dictated by \eqref{eq:sol-lr}. So the set of all possible $\bb{\Theta}$ for random feature maps has zero Lebesgue measure in $\mathbb{R}^{D_r\times (2D+1)}$. It is therefore highly unlikely that gradient descent finds the lower-dimensional subset of the random feature map weights in its search space which is the full $\mathbb{R}^{D_r\times (2D+1)}$. It would be interesting to see if the network generates good features if the loss function is augmented by a penalty term promoting good features. In any case, random feature maps are significantly cheaper to train.
\section{\textcolor{red-}{Results for long-time statistical behaviour}}
\label{sec:longtime}
\begin{figure}[!htp]
     \centering
\includegraphics[scale=0.52]{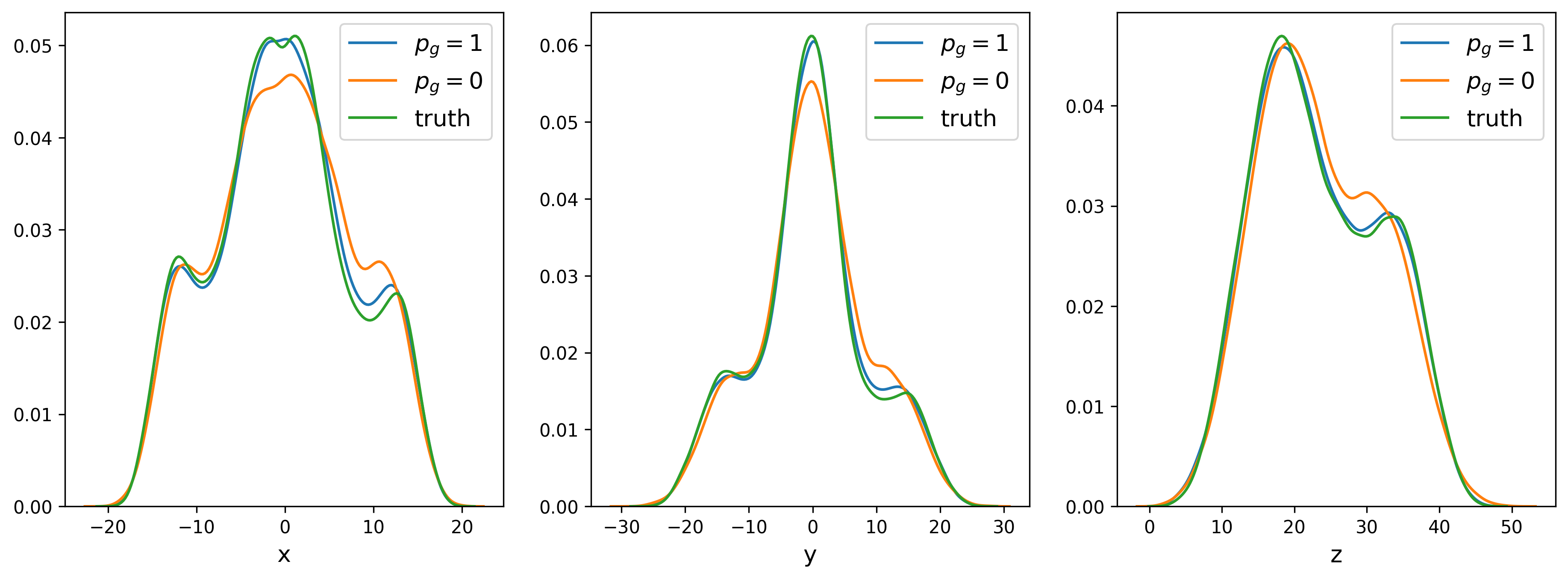}
     \caption{\textcolor{red-}{Empirical histograms of the marginals of the invariant measure for the Lorenz-63 system obtained from simulating the original dynamical system \eqref{eq:L63}, and for random feature map surrogate model with only good features ($p_g=1$) and without any good features ($p_g=0$).}}
     \label{fig:L63hist_pg}
\end{figure}

We have so far focused on short-term integration and assessed the quality of a surrogate model by its ability to remain close to a reference trajectory. In certain applications such as climate prediction, however, it is more important to reliably recover the statistical properties of the dynamical system. Good short-term forecasting, i.e. high mean forecasting times, does not necessarily imply reliable reproduction of the statistics, and vice versa. \textcolor{red-}{In Figure~\ref{fig:L63hist_pg} we compare the empirical histograms of the three variables of the Lorenz-63 system \eqref{eq:L63} with those of the corresponding random feature surrogate maps where we only used good features, i.e. $p_g=1$, and only used linear and saturated features, i.e. $p_g=0$ with $p_l=p_s=0.5$. The histograms are obtained from long simulations over $2,000$ time units, approximating the invariant measure. It is seen that the random feature map with only good features reproduces the invariant measure well. When only linear and saturated features are employed, the histogram of the Lorenz-63 system is less well reproduced. Remarkably, for $p_g=0$ the histogram is still reasonably well reproduced; in particular, the tail behaviour of the histograms is well reproduced.} 

\begin{figure}[!htp]
     \centering
\includegraphics[scale=0.52]{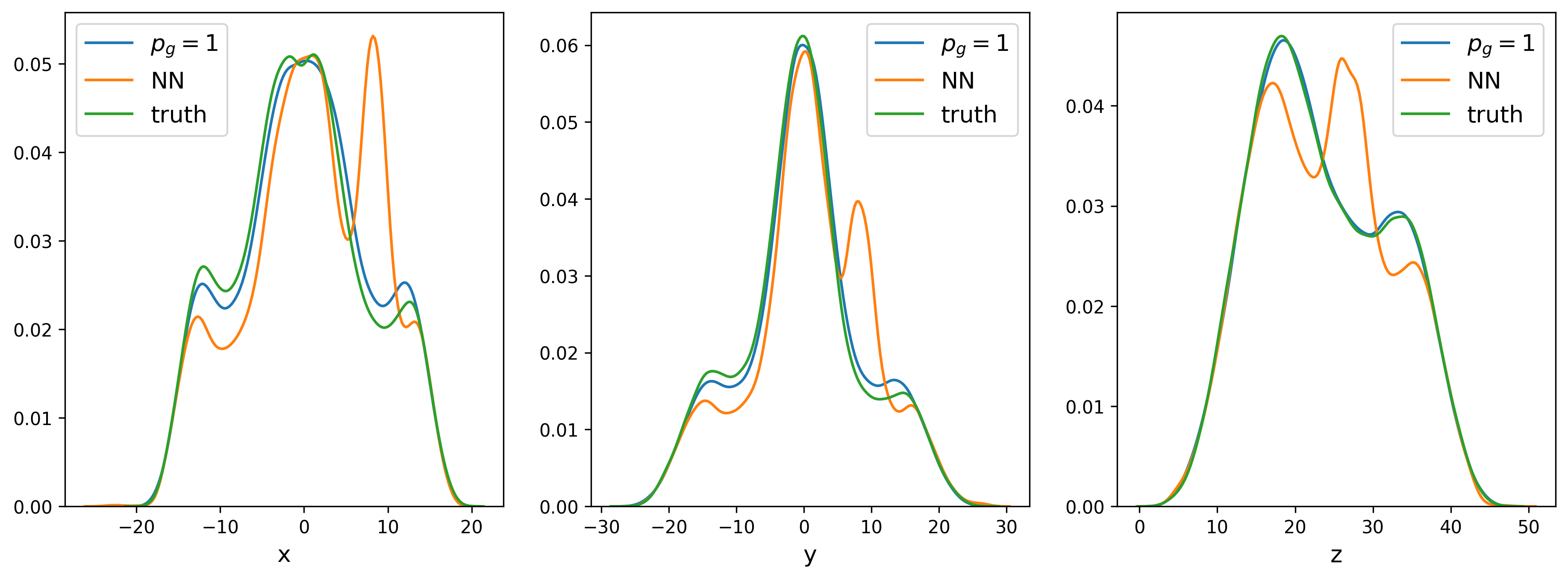}
     \caption{Empirical histograms of the marginals of the invariant measure for the Lorenz-63 system obtained from simulating the original dynamical system \eqref{eq:L63}, for a random feature map surrogate model with $p_g=1$,  and for a neural network (NN) surrogate model. The same neural network is used as in Figure~\ref{fig:nn-evol}.}
     \label{fig:L63hist}
\end{figure}

We further show in Figure~\ref{fig:L63hist} a comparison between the histograms obtained from the random feature map surrogate model that used only good features with $p_g=1$, and the single-layer feedforward network. It is clearly seen that the neural network is not able to reproduce the long-time behaviour, with its empirical histogram being wildly different from that of the original Lorenz-63 system. However, the neural network also reproduces the tail behaviour of the histogram very well.

It is remarkable that including global information about the data, in the form of the constraints \eqref{eq:row-ineq-good} which take into account information about the attractor, is sufficient to ensure that the trained surrogate model is able to recover the invariant measure. This makes random feature models with only good internal weights a very attractive network architecture for dynamical systems. 

%

\section{Summary and future work}
\label{sec:conclude}

We established the notion of good features and good internal parameters for random feature maps with a $\tanh$-activation function. These good internal weights are characterised by affinely mapping the training data into the nonlinear, non-saturated domain of the $\tanh$-activation function. We established that the number of good features $N_g\le D_r$ is the controlling factor in determining the forecasting skill of a learned surrogate map, rather than the feature dimension $D_r$. Interestingly, the forecasting skill was found to be equally deteriorated by linear features as by saturated features.  
We developed computationally cheap hit-and-run sampling algorithms to uniformly sample from the set of good internal parameters. \textcolor{blue-}{The hit-and-run Algorithms~\ref{algo:hr} and \ref{algo:hr-D} sample the internal weights from a data-informed convex set, rather than obtaining them by minimizing some cost function.} We demonstrated how ridge regression engages with a given number of good and bad features. \textcolor{red-}{In particular, saturated features are eliminated almost entirely by the outer weights, provided a sufficient number of good features are present. Similarly, linear features are suppressed by the outer weights, albeit to a lesser degree.} Once there are sufficiently many good features present to allow for a significant reduction of the data mismatch term of the loss function, regularization kicks in and reduces the norm of the outer weights corresponding to good features.\\

We further showed that a single-layer feedforward network with the same width $D_r$ trained with gradient descent exhibits inferior forecasting skill compared to a random feature map surrogate map which used only good internal parameters. The neural network achieves a significantly smaller value of the loss function. Good forecasting  skill, however, requires multiple applications of the surrogate map, and, as we showed, is controlled by the number of good features. The lower forecasting skill is due to the optimization process not finding solutions on the measure-zero set of good parameters. Even when initialized with good parameters, the gradient descent quickly reduces the number of good internal parameters.\\

We found numerically that including global information about the attractor, in form of the constraints for the internal weights \eqref{eq:row-ineq-good}, is sufficient to ensure that the trained random feature map reproduces the statistical properties of the underlying dynamical system and its invariant measure. The rationale for the choice of good internal weights was entirely motivated by the nonlinear non-saturated structure of the $\tanh$-activation function. How far the constraints on the function domain \eqref{eq:row-ineq-good} translate into dynamical information ensuring the preservation of the invariant measure remains an open question, planned for further research.\\

The proposed optimization-free algorithm to choose internal non-trainable parameters can potentially lead to new design and computationally cheap training schemes for more complex network architectures. Our algorithms may be used to further improve the forecasting skill of reservoir computers \cite{pathak2018model,GauthierEtAl21}. \textcolor{red-}{In \cite{MandalGottwald25} it is shown that the hit-and-run algorithm can be used for initializing modified RFMs, which include skip connections, and deep architectures, to achieve state-of-the-art forecasting skill, outperforming standard reservoir computers with significantly less computational effort and model sizes.}\\

\textcolor{blue-}{The parameters $L_0$ and $L_1$ that define {\em{good}} parameters can, in principle, be considered as hyperparameters, and could be tuned, for example, using Bayesian optimization \cite{LevineStuart22}. We have refrained here from doing so as we have not observed significant changes in the forecasting skill for values close to $L_0=0.4$ and $L_1=3.5$ which we used throughout this work.}\\

\textcolor{blue-}{We considered here the $\tanh$-activation function as it is widely used in reservoir computing. The separation of the domain into linear, saturated and good regions can readily be extended to other continuous sigmoidal functions. These activation functions are widely used for random feature maps and in reservoir computing architectures. They are less frequently used for deep neural networks where the saturated regions lead to the vanishing gradient problem \cite{goodfellow2016deep}. However, we note that recently the $\tanh$-function gained again more interest in the machine learning community \cite{ZhuEtl25}. It is an interesting question whether our methodology can be extended to other commonly used activation functions such as the ReLU-type functions or trigonometric functions as initially employed for random feature maps \cite{RahimiRecht08}. 
For example, for GELU~\cite{hendrycks2016gaussian} one may define good features to be those that correspond to an interval around $0$ where the function is nonlinear and non-saturated. For classical Fourier random feature maps, one could separate the domain into linear and saturated regions near the roots and maxima/minima of the $\sin$- and $\cos$-functions, and consider the complement as the ``good" region; the linear domain of the $\sin$-function needs to be aligned with the saturated domain of the $\cos$-function, and vice versa. Exploring whether the good regions of such functions allow for sufficient variability of the corresponding features and if this can lead to an improved sampling is planned for future research.}\\



\section*{Acknowledgement} The authors would like to acknowledge support from the Australian Research Council under Grant No. DP220100931. 


\section{Appendix}
\subsection{Adaptive learning rate for the single-layer neural network}
\label{sec:rate}
We describe in Algorithm~\ref{algo:lr} the adaptive learning rate algorithm we used when training the single-layer feedforward network in Section~\ref{sec:NN}. Essentially our scheduler computes the decay rate of the loss every $I$ steps and modifies the learning rate by increasing or decreasing it by a constant fraction $\xi$ if necessary. We use an initial rate $\eta_0=10^{-3}$,  update interval $I=100$, update fraction $\xi=0.1$, update threshold $\gamma=-10^{-4}$ and number of gradient descent steps $E=1.5\times10^6$ in our scheduler.  

\begin{figure}[!htp]
    \centering
    \includegraphics[scale=0.6]{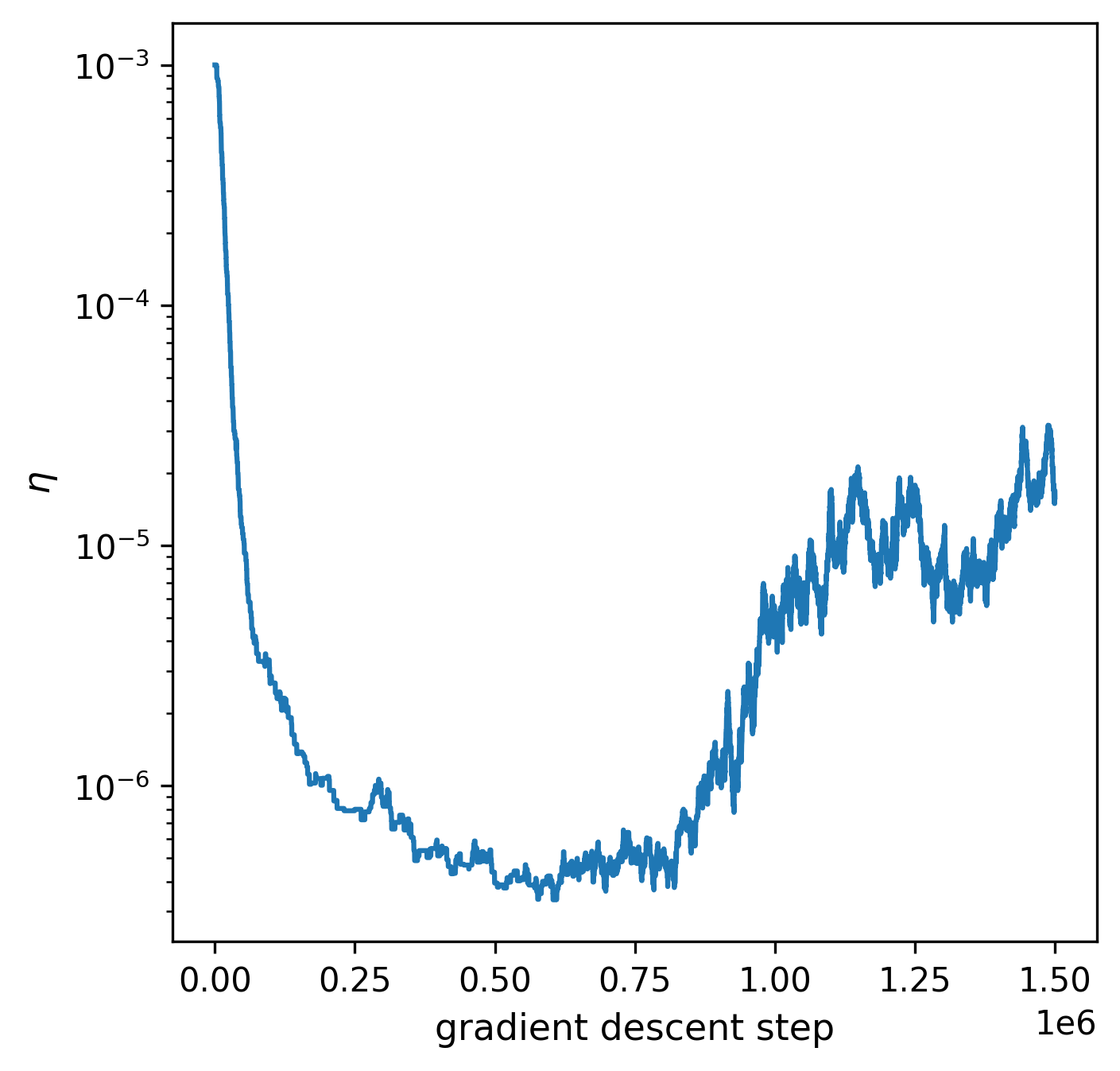}
    \caption{Adaptive learning rate $\eta$ used during the training instance presented in Figure~\ref{fig:nn-evol}, where the loss function is shown.}
    \label{fig:adaptive-lr}
\end{figure}

\begin{algorithm}[!htp]
\caption{Adaptive learning rate scheduler}
\label{algo:lr}
\begin{algorithmic}[1]
\STATE Input: Choose initial rate $\eta_0$, update interval $I$, update fraction $\xi$, update threshold $\gamma$, number of gradient descent steps $E$.
\STATE $k\leftarrow 1$ (gradient descent step).
\STATE $\mathcal{L}_0\leftarrow$ value of $\mathcal L$ at gradient descent step $k$.
\STATE $\eta\leftarrow\eta_0$ (learning rate).
\WHILE{$k<E$}
    \IF{$k$ is divisible by $I$}
        \STATE $\mathcal{L}_1\leftarrow$ value of $\mathcal L$ at gradient descent step $k$.
        \STATE $\Delta\leftarrow\frac{\mathcal{L}_1-\mathcal{L}_0}{\mathcal{L}_0}$.
    \IF{$\Delta > \gamma$}
        \IF{$\Delta >0$}
            \STATE $\eta\leftarrow\eta(1-\xi)$
        \ELSE
        \STATE $\eta\leftarrow\eta(1+\xi)$
        \ENDIF
    \ENDIF
    \STATE $\mathcal{L}_0\leftarrow \mathcal{L}_1$
    \ENDIF
    \STATE$k\leftarrow k+1$
\ENDWHILE
\end{algorithmic}
\end{algorithm}
We tried several other strategies such as finding an optimal learning rate every few steps using bisection, random modifications of the learning rate based on the behavior of the loss function, aggressive constant learning rates, conservative constant learning rates, piecewise linear learning rates etc. We found that the simple strategy presented in Algorithm~\ref{algo:lr} leads to the lowest final value of the loss function for the same number of gradient descent steps. Figure~\ref{fig:adaptive-lr} shows the adaptive learning rate used during the training instance shown in Figure~\ref{fig:nn-evol}. 


\bibliographystyle{plain}
\bibliography{references}

\end{document}